\documentclass[10pt,journal,compsoc]{IEEEtran}
\ifCLASSOPTIONcompsoc
  \usepackage[nocompress]{cite}
\else
  \usepackage{cite}
\fi
\ifCLASSINFOpdf
\else
\fi
\usepackage{amsmath,amssymb,amsfonts,amsthm,mathtools}
\allowdisplaybreaks[4]
\usepackage{array}
\usepackage{url}

\usepackage[english]{babel}
\usepackage{graphicx}
\usepackage{float}
\usepackage{nomencl}
\usepackage{multirow}
\usepackage{lipsum}
\usepackage{algorithm,algorithmic}
\usepackage{epstopdf}
\usepackage{mathrsfs} 
\usepackage{upgreek}
\usepackage{caption}
\usepackage{graphicx}\usepackage{lettrine}
\usepackage{bm,bbm}
\usepackage{lipsum}
\usepackage{nomencl}\usepackage{stfloats}
\usepackage[T1]{fontenc}    
\usepackage{hyperref}       
\usepackage{booktabs}       
\usepackage{amsfonts}       
\usepackage{nicefrac}       
\usepackage{microtype}      
\usepackage{xcolor}         
\usepackage{microtype}
\usepackage{graphicx}
\usepackage{booktabs} 
\usepackage{hyperref}
\usepackage{ragged2e}

\usepackage{color}
\usepackage{amsmath}
\usepackage{amssymb}
\usepackage{mathtools}
\usepackage{amsthm}
\usepackage{amsfonts,extarrows}
\usepackage{bm}
\usepackage{multirow}
\usepackage{subfigure} 

\newtheorem{corollary}{Corollary}
\newtheorem{theorem}{Theorem}
\newtheorem{lemma}{Lemma}

\definecolor{myhlcolor}{rgb}{1, 1, 0}
\definecolor{myhlcolortwo}{rgb}{1, 1, 0}
\definecolor{myhlcolor}{rgb}{1, 1, 1}

\title{Understanding and Improving Model Averaging in Federated Learning on Heterogeneous Data}

\author{Tailin~Zhou, \IEEEmembership{Graduate Student Member,~IEEE,}
        Zehong~Lin, \IEEEmembership{Member,~IEEE,}
        Jun~Zhang,~\IEEEmembership{Fellow,~IEEE,}
        and~Danny~H.K.~Tsang,~\IEEEmembership{Life~Fellow,~IEEE}

\thanks{
This work was supported in part by the Hong Kong Research Grants Council under the Areas of Excellence scheme grant AoE$/$E-601$/$22-R,  in part by NSFC$/$RGC Collaborative Research Scheme grant CRS$\_$HKUST603$/$22,  in part by Guangzhou Municipal Science and Technology Project under Grant 2023A03J0011, in part by Guangdong Provincial Key Laboratory of Integrated Communications, Sensing and Computation for Ubiquitous Internet of Things, and in part by National Foreign Expert Project under Project Number G2022030026.

T. Zhou is with IPO,  Academy of Interdisciplinary Studies, The Hong Kong University of Science and Technology, Clear Water Bay, Hong Kong SAR, China (Email: tzhouaq@connect.ust.hk).
Z. Lin and J. Zhang are with the Department of Electronic and Computer Engineering, The Hong Kong University of Science and Technology, Clear Water Bay, Hong Kong SAR, China  (E-mail: $\{$eezhlin, eejzhang$\}$@ust.hk).
D. H.K. Tsang is with the Internet of Things Thrust, The Hong Kong University of Science and Technology (Guangzhou), Guangzhou, Guangdong, China, and also with the Department of Electronic and Computer Engineering, The Hong Kong University of Science and Technology, Clear Water Bay, Hong Kong SAR, China (Email: eetsang@ust.hk). 
(Corresponding author: Zehong Lin.)
 }
}
 
\begin{document}

\IEEEtitleabstractindextext{%

\justify{  
\begin{abstract}
Model averaging is a widely adopted technique in federated learning (FL) that aggregates multiple client models to obtain a global model. 
Remarkably, model averaging in FL yields a superior global model, even when client models are trained with non-convex objective functions and on heterogeneous local datasets.
However, the rationale behind its success remains poorly understood.
To shed light on this issue, we first visualize the loss landscape of FL over client and global models to illustrate their geometric properties.
The visualization shows that the client models encompass the global model within a common basin, and interestingly, the global model may deviate from the basin's center while still outperforming the client models.
To gain further insights into model averaging in FL, we decompose the expected loss of the global model into five factors related to the client models. 
Specifically, our analysis reveals that the global model loss after early training mainly arises from \textit{i)} the client model's loss on non-overlapping data between client datasets and the global dataset and \textit{ii)} the maximum distance between the global and client models.
Based on the findings from our loss landscape visualization and loss decomposition, we propose utilizing iterative moving averaging (IMA) on the global model at the late training phase to reduce its deviation from the expected minimum, while constraining client exploration to limit the maximum distance between the global and client models.
Our experiments demonstrate that incorporating IMA into existing FL methods significantly improves their accuracy and training speed on various heterogeneous data setups of benchmark datasets.
Code is available at \url{https://github.com/TailinZhou/FedIMA}.
\end{abstract}
} 


 }

 \markboth{To appear in IEEE Transactions on Mobile Computing. 
  Copyright was transferred to IEEE.  DOI: \href{https://doi.org/10.1109/TMC.2024.3406554}{10.1109/TMC.2023.3325366}.}%
{}
 
\maketitle

\IEEEraisesectionheading{\section{Introduction}\label{sec:introduction}}

%
%
%
%

\IEEEPARstart{F}{ederated} learning (FL)  \cite{mcmahan2017communication} enables clients to collaboratively train a machine learning model while keeping their data decentralized to protect privacy.
One of the primary challenges in FL is the heterogeneous data across clients, which diverges client models and deteriorates the performance of FL \cite{zhao2018federated}.
Despite this challenge, numerous works have effectively integrated FL into the artificial intelligence (AI) services of large-scale networks with enormous data to ensure the smooth operation of these networks, including the Internet of Things (IoT) \cite{wu2020fedhome,zhang2021optimizing}, wireless networks \cite{wang2022performance, lin2023cflit}, mobile networks \cite{nguyen2021toward,feng2023feddd} and vehicular networks \cite{zhang2023vehicle}.
The FL empirical success suggests that FL may surpass its theoretical expectations  \cite{kairouz2021advances}.

A common view of the empirical success is that federated model averaging (FMA) mitigates the effect of heterogeneous data in FL, as per \cite{wang2021field}.
Model averaging, first introduced in \cite{polyak1992acceleration},  is a widely used technique to reduce communication overhead \cite{lian2017can} and the variance of gradients \cite{zhang2016parallel}  in distributed/decentralized learning \cite{koloskova2020unified} by periodically averaging models trained over parallel workers with homogeneous data.
In this work, we refer to model averaging in FL on heterogeneous data as FMA to distinguish it from model averaging in other communities (e.g.,  distributed learning on homogeneous data).
Specifically, at each round, FMA aggregates  $K$   client models   updated locally on heterogeneous data to  obtain a global model  $\mathbf{w}$  as 
$\mathbf{w} \leftarrow \sum_{k=1}^K  ({n_k}/{\sum_k n_k})\mathbf{w}_{k}$,
where    $\mathbf{w}_k$ is the $k$-th client model and $n_k$ is  the  size of the $k$-th client dataset.
Surprisingly,   FMA  can work with divergent client models and alleviate their impact on FL  \cite{kairouz2021advances}.

However,  it remains unclear how FMA mitigates the effect of divergent client models and enables the global model to converge throughout the training process.
Existing works, e.g., \cite{li2019convergence,yu2019parallel}, analyze the convergence rate of FMA-based FL under the assumption of bounded gradient dissimilarity.
Specifically, these analyses use an assumed upper bound on the distance between the global and client gradients to ensure a theoretical convergence rate of FL.
Nevertheless, the assumed bound omits the correlations (i.e., covariance) across clients.
A recent work \cite{wang2022unreasonable} demonstrates that the gradient dissimilarity can be arbitrarily large, and the data heterogeneity has no negative impact when ensuring convergence by making this assumption.
Meanwhile, the actual drift of the global gradient (i.e., average client gradients) is significantly smaller than expected based on this bound.
This indicates that the bounded gradient dissimilarity cannot accurately characterize the effect of heterogeneous data on FL.
Since the bound neglects the overall relationship among all client gradients, the effect of FMA on FL is ignored, while FMA plays a practical role in alleviating the drift of the global gradient \cite{wang2021field}.
Therefore,  a conclusive explanation of how FMA assists FL is still lacking.

To fill this gap, we first investigate the geometric properties of FMA by visualizing the loss/error landscape based on the global and client models. 
Our investigation reveals that the global model is closely surrounded by client models within a common basin and consistently achieves lower test loss and error.  
Then, we decompose the expected global model loss to establish a connection between the global model's loss and the client models' losses.
Based on this connection,  we analyze how the global model loss is affected by five factors throughout the training process: training bias, heterogeneous bias, model-prediction variance, covariance between client models, and locality.
Our visualization and decomposition demonstrate that FMA may deviate the global model from the expected center of the loss basin when facing heterogeneous data.
To mitigate the deviation, we employ iterative moving averaging (IMA) on global models along their optimization trajectory.
By integrating IMA into various FL methods,  we can effectively reduce the global model's deviation and keep the model in a low-loss region, thereby improving the training performance.

This work aims to unravel the underlying mechanism of FMA and improve it based on the properties of loss landscape and loss decomposition in FL on heterogeneous data. Unlike previous works on loss landscapes with homogeneous data, this study focuses on unraveling the landscape on FMA with heterogeneous data.
Meanwhile, compared to previous loss decomposition, our analysis decomposes the FMA model loss on heterogeneous data into five factors instead of two. 
This comprehensive decomposition quantifies the performance of the FMA model by analyzing the performance of client models on their respective local datasets.
It is worth noting that our work aims to decompose the global model loss to evaluate the impact of FMA on data heterogeneity in FL, in contrast to \cite{wang2022unreasonable}, which focuses on computing the average client gradient drift at the optimum to ensure a tighter FL convergence rate.
Our main contributions are summarized as follows:
\begin{itemize}
    \item We investigate the dynamics of test loss and classification error landscapes over the global and client models.
    Through these landscape visualizations, we observe that while achieving lower loss/error than client models, the global model is closely surrounded by client models in a common basin but may deviate from its lowest point.
    \item We decompose the global model loss by analyzing the bias and variance of client models on the global dataset. We demonstrate that after early training, the global model loss is dominated by the losses of client models on non-overlapping data between their datasets and the global dataset, as well as their maximum distance from the global model.
    \item Our loss visualization and decomposition indicate that FMA may shift the global model away from the expected point.
    We introduce  IMA  on global models and decay client exploration in late training stages to mitigate this deviation.
    \item  Our experiments show that IMA improves the performance of existing FL methods on various benchmark datasets, enhancing model accuracy and reducing communication costs.
\end{itemize}

   The remainder of this paper is organized as follows.
Section \ref{section:Related Work} reviews related works to ours.
Section \ref{section:Preliminaries} introduces preliminaries on FL and loss landscape visualization. 
The loss landscape of FMA is visualized in Section \ref{section:loss_landscape}, and we present our theoretical and empirical analysis of FMA in Section \ref{section:loss decom}.
Section \ref{section:proposed_method} outlines our proposed method for improving FMA, while simulation results are presented in Section \ref{section:Experiments}.
Finally, the concluding remarks are presented in Section \ref{section:conclusion}.

\section{Related Works}\label{section:Related Work}

 \subsection{Model Averaging}

Model averaging in machine learning (ML) is a technique developed to reduce the variance of model updating by periodically averaging models trained over multiple rounds.  
It was first introduced to average models along the training trajectory in centralized training \cite{polyak1992acceleration}, and then widely adopted to average models over parallel workers in distributed learning \cite{lian2017can,zhang2016parallel,jain2018parallelizing}.
 Izmailov et al. \cite{izmailov2018averaging} discover that a converged ML model tends to end up at the boundary rather than the center of its loss basin while maintaining low loss. 
To encourage convergence to the basin center, they propose stochastic weight averaging (SWA) to average model weights along the optimization path in the final stage.
Notably, SWA does not reinitialize training with the averaged model, thus preserving the optimization trajectory. 
This approach has been extended to distributed learning \cite{gupta2020stochastic} and FL  \cite{caldarola2022improving}.
 Furthermore, maintaining models with mild diversity in the model ensembling and model average \cite{lee2016stochastic,rame2022diverse} has improved the model generalization.

A comprehensive survey \cite{kairouz2021advances} indicates that although FMA has achieved empirical success in FL,  its underlying mechanisms remain unclear.
Unlike traditional model averaging on homogeneous data, FMA  needs to accommodate the challenges posed by heterogeneous data in FL \cite{mcmahan2017communication}.
Notably,  despite clients optimizing non-convex  ML objectives on heterogeneous local datasets, FMA consistently achieves a converged global model by aggregating divergent client models.
Therefore, to understand the mechanism of FMA, we begin by analyzing its geometric properties on heterogeneous data through loss landscape visualization, followed by the decomposition of the expected loss.
  
\subsection{FL on Heterogeneous Data}

Heterogeneous data across clients is one of the primary challenges in FL  \cite{zhao2018federated}. 
 Common solutions involve improving the local training on clients or modifying model aggregation on the server.
Client-side methods typically introduce regularization to local loss functions to prevent local models from converging to their local minima instead of the FL minima.
For example, regularization can be designed as the distance between client and global models in FedProx \cite{li2020federated}, the distance between feature anchors and features in FedFA \cite{zhou2022fedfa}, or the distance among client-invariant features in FedCiR \cite{li2023fedcir}.
These approaches aim to handle heterogeneous data on the client side, but they do not improve FMA.
On the other hand, server-side methods develop alternative aggregation schemes building upon FMA.
For instance, before performing FMA at the server, FedNova \cite{wang2020tackling} normalizes local updates to mitigate the impact of varying numbers of local updates, FedAdam and FedYogi \cite{reddi2021adaptive}  introduce adaptive momentum to mitigate updating oscillation of the global model, 
and FedGMA \cite{tenison2022gradient} applies the AND-Masked gradient update to sparsify the global model and improve the loss flatness. Moreover, some methods allow clients to tackle heterogeneous data by sharing data with privacy guarantees, such as sharing synthesized \cite{li2022federated} or coded data \cite{sun2022stochastic,shao2022DReS}.
In addition to addressing heterogeneous data, several studies have explored techniques, including specification \cite{PandaMBCM22}, quantization  \cite{ShlezingerCEPC21}, and low-rank decomposition \cite{lan2023communication}, to reduce the communication overhead from a global model perspective.

While improving FL performance on heterogeneous data, analyses of existing works like \cite{yu2019parallel,li2020federated,karimireddy2020scaffold,wang2020tackling}  mainly focus on the overall convergence of their proposed methods, rather than elucidating the success of FMA.
A common assumption of these analyses is the bounded dissimilarity of client gradients \cite{li2019convergence,wang2020tackling}, but this assumption is overly pessimistic, as indicated in \cite{wang2022unreasonable}. 
It fails to characterize the practical drift of the global model, which is much smaller after FMA on client updates than the theoretical expectation.
As suggested in \cite{wang2021field},    FMA may maintain the drift close to zero on heterogeneous data, though the underlying rationale remains unclear.
To fill this gap, we investigate how FMA achieves success and how to improve it during the training.

\subsection{Loss Landscape Analysis}
Loss landscape \cite{sun2020global} analysis refers to the visualization and understanding of the optimization landscape of a model's loss function.
It is a common approach to provide insights into models' convergence, generalization, and geometric properties.
The loss landscape is typically visualized by plotting the loss function through low-dimensional projections along random or meaningful directions in the parameter space  \cite{goodfellow2014qualitatively,izmailov2018averaging,li2018visualizing}.
In \cite{goodfellow2014qualitatively}, Goodfellow et al. take the first step to visualize the optimization trajectory of ML models using low-dimensional projections, enabling comparison of different optimization algorithms.
In \cite{li2018visualizing}, sharp and flat minima concepts are introduced, where flat minima generally yield better generalization. 
Subsequently,   Izmailov et al. \cite{izmailov2018averaging} leverage landscape visualization to show that a converged ML model tends to end up at the boundary of the wide flat region (i.e., a loss basin)  instead of its flatter center.
In addition,  sharpness aware minimization (SAM) is introduced to seek flat minima \cite{foret2020sharpness} and extended to domain generalization \cite{cha2021swad} and FL \cite{caldarola2022improving}.
Regarding geometric structure, Garipov et al. \cite{garipov2018loss} discover that different minima in ML models have a connected structure called mode connectivity despite facing non-convex challenges.
This implies that local minima are not isolated but interconnected within a manifold \cite{draxler18Essentially}.

While loss landscape visualization has been extensively studied, previous research has primarily focused on centralized training with homogeneous data.
For instance, the researchers of \cite{izmailov2018averaging,sun2020global} only consider data shuffling, where the data distribution remains consistent across all workers.
In distributed learning, which is more similar to FL,  some studies like \cite{zhang2016parallel} have explored when model averaging helps distributed training, suggesting that model averaging brings models of different workers to a common basin of attraction. However, these studies do not provide further visualization analysis and only consider the workers' data independently drawn from the same data pool, i.e., homogeneous data.

In contrast, our work specifically addresses FL scenarios, which have received comparatively less attention in terms of loss landscape visualization.
A few preliminary studies \cite{caldarola2022improving,li2022understanding,li2023revisiting,zhou2023mode} have attempted to visualize the loss landscape and improve FL performance by enhancing loss flatness. 
Nonetheless, these studies have not directly analyzed how FMA handles data heterogeneity for FL, nor have they explored the bias introduced by the global model in the landscape.
In contrast, our work stands out by delving into the geometric properties of FMA to understand its underlying mechanism and provide a clear visualization of its loss landscape.
Specifically, our work aims to fill this gap by investigating how FMA enables the convergence of the global model aggregated by client models trained on heterogeneous data, despite using a non-convex objective function.
More importantly, we demonstrate a novel finding that the global model may deviate from the expected point when using FMA. 
To the best of our knowledge, our study is the first to identify the deviation of global models from the basin's center on the landscape.

\subsection{Bias-variance Loss Decomposition}
Bias-variance loss decomposition is a helpful concept for understanding the performance of ML models \cite{geman1992neural,kohavi1996bias,domingos2000unified,brown2005between}. 
Specifically, the expected model loss is decomposed into bias, variance, and irreducible error components.
Bias quantifies the fitting capability of models on the training data, while variance reflects the models' sensitivity to small fluctuations in training data \cite{geman1992neural}.
In \cite{domingos2000unified}, a unified bias-variance decomposition framework for regression and classification tasks is proposed to guide model selection in the model ensembling. 
 Belkin et al. \cite{belkin2019reconciling} study how deep ML models achieve low bias and variance via this decomposition.

Compared to previous bias-variance decomposition works that focus on homogeneous data, our analysis delves into heterogeneous data and further decomposes the bias and variance into five factors: training bias, heterogeneous bias, variance, covariance, and locality. 
This novel decomposition aims to elucidate the underlying mechanism of the global model on heterogeneous data.
Specifically, we decompose the bias factor into training bias and heterogeneous bias, and re-derive the variance as the variance and covariance factors among client models that are not independent and identically distributed.
In addition, we introduce a locality factor to control the effectiveness of our decomposition by measuring the maximal distance between the client and global models.
This comprehensive decomposition enables us to identify the main factors that affect the global model when FL performs FMA on heterogeneous data.

Moreover, it is important to note that our decomposition differs from bias/variance-reduction optimization techniques, such as FedADAM/FedYogi \cite{reddi2021adaptive} and Scaffold \cite{karimireddy2020scaffold}, which aim to control the bias/variance of gradient updates to accelerate convergence.
Our primary focus is to characterize the performance of the global model on the global dataset by decomposing the performance of client models on their respective local datasets, which supports our geometric observation of the loss landscape on FMA. 
The different motivation behind the decomposition allows our proposed IMA approach to complement and enhance these variance-reducing methods rather than conflict with them.

\section{Preliminaries}\label{section:Preliminaries}

\subsection{Federated Learning (FL)}
\subsubsection{FL Problem Formulation}
We consider an FL framework with $K$ clients,   each possessing its dataset $\mathcal{D}_k = \{(x,y)\}  \sim \mathcal{P}_k $ consisting of $n_k$ data samples, where $x $ and $y$ denote a labeled data sample and its corresponding label, respectively.
The global dataset of FL is the union of all client datasets and denoted by $\mathcal{D} = \cup_{k=1}^K \mathcal{D}_k \sim \mathcal{P}$, comprising  $n = \sum_{k=1}^N n_k$ data samples.
Here, $ \mathcal{P}_k $ and $ \mathcal{P}$ represent the client and global data distribution, respectively.
When dealing with a ML task on the global dataset $\mathcal{D}$, FL uses a finite-sum objective to minimize the expected global loss $  \mathcal{L}(\mathbf{w}):= \mathbb{E}_{(x, y) \in \mathcal{D}}[  l (\mathbf{w};(x, y))]$,   where $l(\mathbf{w})$   denotes the global loss function for model parameters $\mathbf{w}$.  
As shown in \cite{mcmahan2017communication}, this objective  can be reformulated as: 
\begin{equation}
 \begin{aligned}
   \min_{\mathbf{w} \in \mathbb{R}}  \mathcal{L}(\mathbf{w} )  = &    \sum_{k=1}^K \frac{n_k}{n} 
 \mathcal{L}_k(\mathbf{w} )  
    =     \sum_{k=1}^K \frac{n_k}{n} \sum_{i=1}^{n_k}  l_k (\mathbf{w};({x}_i, y_i)\in  \mathcal{D}_k),
    \end{aligned}
   \label{fl}
\end{equation}
where  $\mathcal{L}_k(\cdot)$ is the expected local loss  of the $k$-th client on its local dataset $\mathcal{D}_k$, and  $l_k(\mathbf{w})$  is the local loss function on $\mathbf{w}$. 
 
An FL method called FedAvg \cite{mcmahan2017communication} optimizes the objective (\ref{fl}) by averaging client models at the server in a periodic manner. In each round, the method has the following steps:
\begin{enumerate}
    \item Clients update their local models  $\{\mathbf{w}_k\}_{k=1}^K$ independently by minimizing their local losses $\{ \mathcal{L}_k (\mathbf{w}_k) \}_{k=1}^K$ on the local datasets $\{ \mathcal{D}_k \}_{k=1}^K$;
    \item Clients upload their updated models to the server;
    \item The server   performs FMA on the local models  to calculate the new global model, i.e., $\mathbf{w} =\sum_{k=1}^{K} \frac{n_k}{n}\mathbf{w}_k$;
    \item  The new global model $\mathbf{w}$ is sent back to clients to initialize the next round of local training.
\end{enumerate}
This process repeats until the global model converges.

\subsubsection{Heterogeneous Data Problem and FMA in FL}
The objective (\ref{fl}) assumes that the client data distribution $\mathcal{P}_k$ is formed by uniformly and randomly distributing the training examples from the global data distribution $\mathcal{P}$.
 However,  the assumption does not generally hold in FL due to heterogeneous data among clients, where $\mathcal{P}_k   \neq  \mathcal{P}_{k^\prime} \neq \mathcal{P}$ when $k \neq k^\prime$.
As per \cite{zhao2018federated,wang2020tackling,li2020federated}, FL performance can be negatively impacted by heterogeneous data, leading to a slower convergence speed and worse model generalization.
 
There are two common types of heterogeneous data  \cite{kairouz2021advances}: feature distribution skew and label distribution skew, referred to as \textit{feature skew} and \textit{label skew}, respectively, in this work for brevity.
Our work delves into the effect of these two types of heterogeneous data on FMA in FL.
 Suppose that the $k$-th client data distribution follows $\mathcal{P}_k({x},y)=\mathcal{P}_k({x}|y)\mathcal{P}_k(y) =\mathcal{P}_k(y|{x})\mathcal{P}_k({x})$, where $\mathcal{P}_k({x})$ and $\mathcal{P}_k(y)$ denote the input feature marginal distribution and label marginal distribution of the $k$-th client, respectively.
 Specifically,   label skew means that $\mathcal{P}_k(y)$  varies from $\mathcal{P}_{k^\prime}({y})$  while $\mathcal{P}_k({x}|y) = \mathcal{P}_{k^\prime}({x}|y) $ for   clients   $k \neq k^\prime$;
  feature skew  means that  $\mathcal{P}_k({x})$  varies from $\mathcal{P}_{k^\prime}({x})$  while $\mathcal{P}_k(y|{x}) = \mathcal{P}_{k^\prime}(y|{x}) $ for clients $k \neq k^\prime$. 

In FL, when optimizing the objective (\ref{fl})  on heterogeneous data,  $\mathcal{L}_k$ can be an arbitrarily poor approximation to $\mathcal{L}$ \cite{mcmahan2017communication}, e.g., an inconsistent local objective with the FL objective, potentially hindering the FL convergence.
Nonetheless,   FL typically outperforms its theoretical convergence expectation despite data heterogeneity \cite{wang2022unreasonable}.  
For example, FedAvg shows empirical success as per \cite{kairouz2021advances}, with FMA  keeping the global model converging throughout the training process.
A recent survey \cite{shao2023survey} discovers that   FMA effectively balances sharing information among clients while preserving privacy.
This highlights the crucial role of FMA in FedAvg, while it remains unclear how FMA deals with heterogeneous data on FL.

\begin{figure*}[th]
    \centering
    \includegraphics[width=\textwidth]{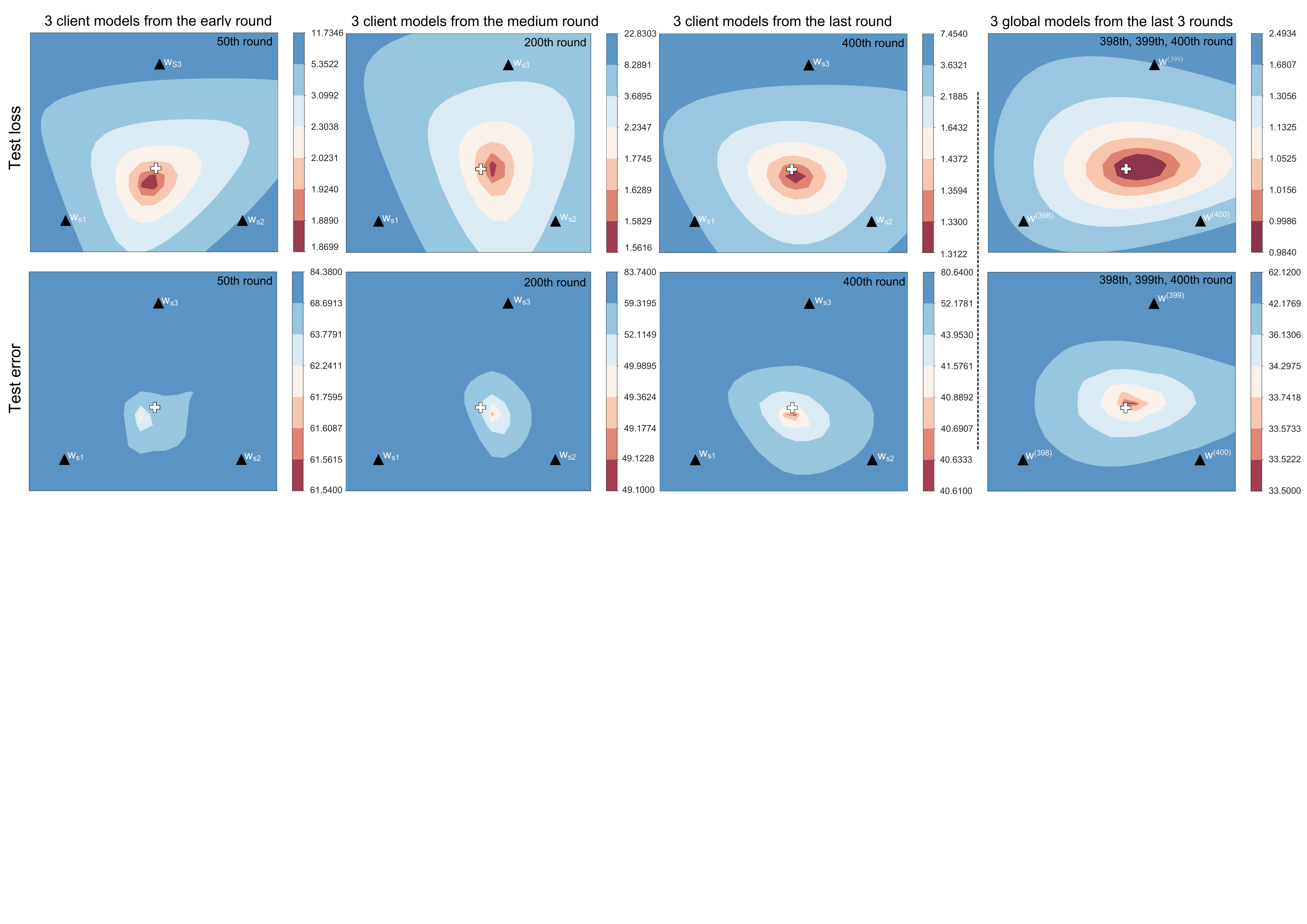}
    \caption{Visualization of the loss  (\textbf{top row}) and classification error   (\textbf{bottom row}) landscapes on the CIFAR-10  test dataset, along with three client models from the early stage (\textbf{first column}), middle stage (\textbf{second column}), and final stage (\textbf{third column}),  as well as the visualization of three global models from the final three rounds (\textbf{fourth column}).
The \textit{black triangles} represent the location of three models in the plane, while the \textit{white cross} represents their average model's location. 
The loss/error landscape can be viewed as a basin, where client models reach the basin's wall and the global model approaches the basin's bottom as FL training proceeds.
FMA helps move the global model towards the basin's bottom by averaging client models on the basin's wall, while heterogeneous data deviates the global model from the basin's center.}

    \label{loss_error_visualization}
\end{figure*}

 \subsection{Loss Landscape Visualization}
The loss landscape depicts the distribution of loss values throughout the model's weight space. 
As per \cite{sun2020global}, exploring the loss landscape can enhance our understanding of ML problems.
While it is generally difficult to visualize the landscape in high-dimensional spaces, there have been many attempts to achieve it by dimensionality reduction.
This helps reveal the geometric properties of neural networks, such as flatness \cite{garipov2018loss} and optimization trajectory \cite{goodfellow2014qualitatively}.
In this work, we employ two common approaches to visualize the loss landscape of FL: 1D and 2D  visualizations.

For 1D visualization, we follow \cite{goodfellow2014qualitatively} to draw the loss landscape in a line segment (1D) by using linear interpolation between two
models $\mathbf{w}_1$ and $\mathbf{w}_2$.
Specifically, given a target dataset, we evaluate the loss of different model weights along the line segment between $\mathbf{w}_1$ and $\mathbf{w}_2$, i.e., $\mathcal{L}_{[\mathbf{w}_1,\mathbf{w}_2]}(\beta)= \mathcal{L}(\beta\mathbf{w}_1+(1-\beta)\mathbf{w}_2)$, where $\beta$ is the interpolation coefficient of line model interpolation between $\mathbf{w}_1$ and $\mathbf{w}_2$.

For 2D visualization,  we explore the loss landscape in a plane (2D) by drawing the contour based on three models according to \cite{izmailov2018averaging}.
Specifically, we take $\mathbf{w}_1,\mathbf{w}_2,\mathbf{w}_3$ to form a plane by constructing  two base vectors
$\mathbf{\bar{u}}=\mathbf{u}/\|\mathbf{u}\|^2 $ and $\mathbf{\bar{v}}=\mathbf{v}/\|\mathbf{v}\|^2 $, where $\mathbf{u} =\mathbf{w}_2- \mathbf{w}_1$ and  
$\mathbf{v}= (\mathbf{w}_3- \mathbf{w}_1) - \left \langle \mathbf{w}_3 - \mathbf{w}_1 , \mathbf{w}_2- \mathbf{w}_1 \right \rangle/\| \mathbf{w}_2- \mathbf{w}_1\|^2 \cdot (\mathbf{w}_2- \mathbf{w}_1)$.
Next, each plane point \textit{P},   with coordinates $(a, b)$,  represents a model $ \mathbf{w}_{\textit{P}} = \mathbf{w}_1 + a\cdot\mathbf{\bar{u}} +  b\cdot\mathbf{\bar{v}} $.
Finally, given a dataset, we evaluate the loss $ \mathcal{L}(\mathbf{w}_{\textit{P}})$ of all the points in this plane and draw the loss contour by $ \{\mathcal{L}(\mathbf{w}_{\textit{P}})=c\}$ with a contour value $c$.

\section{Loss Landscape Visualization in FL}\label{section:loss_landscape}
  In this section,  we explore the geometric properties of FMA through 2D loss landscape visualization.
  As depicted in Figure \ref{loss_error_visualization},   to construct the plane of 2D visualization, we use three client models from the same training round in the first three columns and three global models from different training rounds in the fourth column.
The  FL setup related to Figure \ref{loss_error_visualization} involves training a global model on the CIFAR-10  dataset \cite{krizhevsky2009learning} across 100 clients over 400 rounds.
To introduce data heterogeneity, each client dataset contains two class shards of CIFAR-10, following \cite{mcmahan2017communication} (see specific setups in Table  \ref{Setup_all_visuz_exps}).
In this setup, the ideal accuracy of client models on the CIFAR-10  test set is $20\%$ if clients train their models by local datasets. 
More results for various setups, including different datasets, data heterogeneity, models, and FL settings, are provided in Figures \ref{loss_visualization_cross_silo_FL}, \ref{error_landscape}, and \ref{full_loss_landscape} of the Appendix.

\subsection{Lower Test Loss with FMA}

In Figure \ref{loss_error_visualization}, we observe that the averaged model (i.e.,  the  \textit{white cross}) of the three client models (i.e.,  the \textit{black triangles})  is consistently located at the regions with lower test loss and classification error than individual client models.
This implies that FMA can effectively aggregate local client information into the global model.
Furthermore, the first three columns of Figure \ref{loss_error_visualization} correspond to three different training stages.
As training progresses, since the newly-aggregated global model re-initiates client models with lower loss, FMA prevents them from over-fitting to their respective datasets.
Meanwhile, FMA leverages client models with lower losses to anchor the global model more precisely in a lower-loss landscape area.
In other words, FMA prevents over-fitting information from aggregating into the global model.

Moreover, we observe a bias between the \textit{white cross} and the lowest loss/error point in the first three columns of Figure \ref{loss_error_visualization}.
This bias can be caused by the deviation between the averaged models (i.e., the \textit{white cross}) and the global model or between the global model and its optimal model.
To further investigate this bias, we visualize the loss/error landscape of global models obtained from the final three rounds (i.e., the 398th, 399th, and 400th rounds) in the fourth column of Figure \ref{loss_error_visualization}.
There is a bias between the global models  (i.e., the \textit{black triangle}) and the lowest loss point in the loss landscape, similar to the bias observed over client models, and their averaged model (i.e., the \textit{white cross}) is closer to the lowest point. 
This reveals that the only performing FMA on client models may fail to achieve the optimal global model.
In summary, FMA helps move the global model towards the center of the loss basin during the FL training process.
However, while the global model is converging,   the presence of heterogeneous data causes the global model's movement to deviate from the basin's center.
In Section \ref{section:proposed_method}, we will address the deviation of the global model aggregated by FMA from the lowest loss point.

\subsection{Global Model and Client Models in a Common Basin}
In Figure \ref{loss_error_visualization}, the second row demonstrates that the test classification errors of client models are around $80\%$.
These errors almost reach the lowest classification error obtained by client models through local training, indicating their proximity to local optima. 
Moreover, Figure \ref{loss_error_visualization} illustrates that the averaged model is surrounded by client models and located near a local optimum of the global model throughout the entire training process.
Meanwhile, the distance between global and client models remains limited, as presented in Figure \ref{locality_validation}.
These observations suggest that client models within a common basin closely surround the global model.

Geometrically, the test loss/error landscape in FL can be viewed as a basin, with client models reaching the basin's wall and the global model near the basin's bottom, as shown in Figure \ref{loss_error_visualization}.
This geometric property provides a novel insight into the mechanism behind FMA in FL.
For example,  Wang et al. \cite{wang2022unreasonable} have empirically found that the client-update drifts' practical impact on the global model's convergence speed is less than predicted by theoretical analysis.
This observation can be explained by the geometric property as follows:
In the earlier stages of training, clients update their models towards the basin, resulting in client updates that are roughly in the same direction but not identical. Consequently, the client-update drifts are small in the earlier stages.
However, as client models approach their optimal points in the later stages, they encounter the basin walls in the loss landscape.
That is, clients' optimal points are scattered around the basin's wall.
These client models are then initialized by the global model, which is near the basin's bottom, and the direction of client updates may radiate in all directions, from the basin's bottom to the wall.
Fortunately, when FL performs FMA on the drifts of client updates, these drifts tend to cancel each other out, resulting in a limited impact on the global model and preventing it from drifting away from the basin's bottom.
Therefore, the client-update drifts remain small even in the later stages of training, although their updated directions may be more dissimilar compared to earlier stages.

\section{Expected Loss Decomposition}\label{section:loss decom}
 In this section,   we will analyze the relationship of the losses between the global and client models when FL performs FMA. 
To decompose the expected global model loss, we first examine the connection between FMA and the weighted-model ensembling (WENS).
Next, we decompose the global model's expected loss using this connection based on the client models' losses.
Finally, we empirically validate our decomposition analysis to show which factors dominate the global model's loss throughout the training process.

We represent  the forward function of $\mathbf{w}$ as  $f_{\mathbf{w}}: \mathcal{X} \rightarrow \mathcal{Y}$, where $\mathcal{X}$ and $\mathcal{Y}$ are input and output spaces, respectively. 
For simplicity, we focus on the mean-square error (MSE) loss in the theoretical analysis, i.e., $l (\mathbf{w};(x, y))= (y - f_{\mathbf{w}}(x))^2$. 
It is worth noting that this framework can be extended to other loss functions \cite{domingos2000unified}.  
Due to mode connectivity of neural networks \cite{izmailov2018averaging, draxler18Essentially, zhou2023mode}, given a model architecture $\mathcal{W} \subset \mathbb{R}^d$, a loss function $\mathcal{L}$, and a training dataset $\mathcal{D}_{\rm tr}$,
there exists a single connected low-loss manifold that contains all the minima trained on $\mathcal{D}_{\rm tr}$.
In other words, there exists a model solution subspace $ \mathcal{W}_{\mathcal{D}_{\rm tr}} = \{\mathbf{w}_{\rm tr}\}\subset \mathcal{W}$, where $\mathbf{w}_{\rm tr}$ denotes a model  optimized on $\mathcal{D}_{\rm tr}$.
When the model $\mathbf{w}$ is uniformly distributed  in $\mathcal{W}_{\mathcal{D}_{\rm tr}}$,  the bias-variance decomposition  of  the expected loss of $\mathbf{w}$  evaluated on a test dataset $\mathcal{D}_{\rm te}$ can be expressed as \cite{brown2005between, kohavi1996bias}:
\begin{equation}
\begin{aligned}
      \mathbb{E}_{ \mathbf{w}\in \mathcal{W}_{\mathcal{D}_{\rm tr}}} \mathcal{L}_{\mathcal{D}_{\rm te}} (\mathbf{w})
   =  & \mathbb{E}_{(x,y) \in \mathcal{D}_{\rm te}}[ 
  {\underbrace{\left(y- \bar{f}_{\mathcal{D}_{\rm tr}}(x) \right)}_{{\rm{Bias}} \{f|(x,y)\}}} ^2
 \\  & + \underbrace{\mathbb{E}_{ \mathbf{w}\in \mathcal{W}_{\mathcal{D}_{\rm tr}}} [  \left(f_{\mathbf{w}}(x) - \bar{f}_{\mathcal{D}_{\rm tr}}(x)  \right)^2]}_{{\rm Var}\{f| x\}}
  ],
\end{aligned}
\label{decomposistion:step1}
\end{equation}
 where $f_{\mathbf{w}}(\cdot)$ and $\bar{f}_{\mathcal{D}_{\rm tr}}(\cdot) =  \mathbb{E}_{ \mathbf{w}\in \mathcal{W}_{\mathcal{D}_{\rm tr}}}[f_{\mathbf{w}}(\cdot) ]$ are the model output of $\mathbf{w}$ and  the expected output on $\mathcal{W}_{\mathcal{D}_{\rm tr}}$, respectively.
 
Since $\bar{f}_{\mathcal{D}_{\rm tr}}(\cdot)$  represents the ensemble output of all models in $\mathcal{W}_{\mathcal{D}_{\rm tr}}$, we rewrite  it  as  a finite-sum formulation, $\bar{f}_{\mathcal{D}_{\rm tr}}(\cdot) = \frac{1}{N} \sum_{i\in [N]} f_{\mathbf{w}_i\in \mathcal{W}_{\mathcal{D}_{\rm tr}}}(\cdot)$.
  Specifically, given $ \mathcal{W}_{\mathcal{D}_{\rm tr}}$ and a sample $(x,y) \in \mathcal{D}_{\rm te}$,
 ${\rm{Bias}} \{f|(x,y)\}$    denotes the   bias  between the ground truth $y$   and the ensemble output $\bar{f}_{\mathcal{D}_{\rm tr}}(x)$  and ${\rm Var}\{f| x\}$ denotes the expected MSE  between $f_{\mathbf{w}}(x)$   and   $\bar{f}_{\mathcal{D}_{\rm tr}}(x)$,  which depends on the discrepancy between $\mathcal{D}_{\rm tr}$ and $\mathcal{D}_{\rm te}$ according to \cite{ye2022ood}.
Note that the bias captures the capability of the models to fit the training data distribution $\mathcal{D}_{\rm tr}$, while the variance measures the models' sensitivity to small fluctuations in $\mathcal{D}_{\rm tr}$.

 \subsection{Connection between FMA and WENS}  
At each round, FMA performs weighted averaging, defined as $\mathbf{w}_{\rm FMA} \leftarrow \sum_{k=1}^K \frac{n_k}{n} \mathbf{w}_{k}$,  where the averaging weight depends on $n_k$ \cite{mcmahan2017communication}  and $n= \sum_k n_k$.
According to  \cite{izmailov2018averaging}, the model average is a first-order approximation of the model ensembling when the averaged models are closely located in the weight space, where the model ensembling represents the averaging of outputs from multiple diverse models given the same input. 
Based on this approximation, we establish the connection between FMA and WENS as follows:
\begin{lemma} (FMA and WENS. See proof in Appendix) Given $K$  models $\mathbf{\{w}_k\}_{k=1}^K$ and $n_i/n_j \neq \infty$ when $i\neq j$, we denote  $\Delta_k =\|\mathbf{w}_{k} -\mathbf{w}_{\rm FMA}  \|$ and $\Delta=\max_{k=1}^K\Delta_k$. Then, we have:
  \begin{equation*}
\begin{aligned}
f_{\rm WENS}(x)  - f_{\mathbf{w}_{\rm FMA}}(x)  =  \langle \Delta f_{\mathbf{w}_{\rm FMA}}(x),  \sum_{k=1}^K \frac{n_k}{n}  \Delta_k \rangle +  O(\Delta^2 ),
\label{equation_lemma1}
\end{aligned}
\end{equation*}
where the WENS on the $K$ models is to conduct weighted averaging on the outputs of these models when given the same input, represented as $f_{\rm WENS}(x)= \sum_{k=1}^K \frac{n_k}{n} f_{\mathbf{w}_k}(x)$.
\label{lemma1}
\end{lemma}

 Lemma \ref{lemma1} shows that the output of the FMA model $f_{\rm FMA}(\cdot)$, i.e., the global model in FL, is a first-order approximation of weighted averaging on the outputs of client models, i.e., the WENS $f_{\rm WENS}(\cdot)$.
The term $O(\Delta^2 )$ measures the quadratic of the maximum distance between the client and global models and controls the approximation error.
With a limited maximum distance, the approximation error is expected to be small, which will be verified in Figure  \ref{locality_validation}.
Note that WENS involves averaging model outputs, while FMA involves averaging model parameters.
 The connection between FMA and WEN enables us to conduct a bias-variance decomposition on FMA using equation (\ref{decomposistion:step1}), which relates to model outputs.

\subsection{Expected Loss Decomposition of Global Model}
With Lemma \ref{lemma1}, we can incorporate FMA and adapt the bias-variance decomposition (\ref{decomposistion:step1}) to the FL version.
Specifically, the model $\mathbf{w}$ in (\ref{decomposistion:step1}) is substituted by $K$ client models $\mathbf{\{w}_k\}_{k=1}^K$,   $\mathcal{D}_{\rm tr}$ and  $\mathcal{D}_{\rm te}$ are modified to   client datasets $\{\mathcal{D}_{k}\}_{k=1}^K$ and  the global dataset $\mathcal{D}$, respectively.
Meanwhile, 
  $\bar{f}_{\mathcal{D}_{\rm tr}}(\cdot) =  \mathbb{E}_{\mathbf{\{w}_k\}_{k=1}^K \in \prod_{k}^K \mathcal{W}_{\mathcal{D}_{k}}}[f_{\mathbf{\{w}_k\}_{k=1}^K}(\cdot) ]=\mathbb{E}_{\mathbf{\{w}_k\}_{k=1}^K}[f_{\rm WENS}(\cdot) ]$ denotes the ensemble output of the combination subspace on $K$ client models, where $\prod_{k}^K \mathcal{W}_{\mathcal{D}_{k}} = \mathcal{W}_{\mathcal{D}_{1}} \times \cdots  \times  \mathcal{W}_{\mathcal{D}_{K}}$.
Then, we decompose the expected loss of $f_{\mathbf{w}_{\rm FMA}}(\cdot)$ on $\mathcal{D}$ in the following theorem:

\begin{theorem} (Loss decomposition of the global model. See proof in Appendix) 
Given $K$ client models $ \mathbf{\{w}_k\}_{k=1}^K \in \prod_{k}^K \mathcal{W}_{\mathcal{D}_{k}}$,  the expected loss of the global model $\mathbf{w}_{\rm FMA}$ on  $\mathcal{D}$ is decomposed as:
\begin{equation*}
    \begin{aligned}
  \mathbb{E}_{ \mathbf{\{w}_k\}_{k=1}^K} & \mathcal{L} (\mathbf{{w}_{\rm FMA}} ) 
  = \frac{1}{n} \sum_{(x,y)\in\mathcal{D}}
  [\sum_{k=1}^K \frac{n_k}{n} {\rm{TrainBias}} \{f_{\mathbf{w}_k}|(x,y)\} 
  \\  +&  \frac{n_k}{n}{\rm{HeterBias}} \{f_{\mathbf{w}_k}|(x,y)\} ]^2
  + { \sum_{k=1}^K \frac{n_k^2}{n^2} {\rm Var}\{f_{\mathbf{w}_k}|x\}} 
 \\  +&    {\sum_k \sum_{k^\prime\neq k} \frac{n_k n_{k^\prime}}{n^2}  {\rm Cov}\{f_{\mathbf{w}_k,\mathbf{w}_k^\prime}|x\}}  
  +  O(\Delta^2 ),
    \end{aligned}
    \label{equation of theorem1}
\end{equation*}
where 
${\rm{TrainBias}} \{f_{\mathbf{w}_k}|(x,y)\} =  \mathbb{I}[{(x,y)\in\mathcal{D}_k}]  (y- \mathbb{E}_{ \mathbf{w_k}} [  f_{\mathbf{w}_k}(x)] )$; 
${\rm{HeterBias}} \{f_{\mathbf{w}_k}|(x,y)\}  =   \mathbb{I}\left[{(x,y)\in \mathcal{D} \setminus \mathcal{D}_k}\right] $ $  (y- \mathbb{E}_{ \mathbf{w_k}} [  f_{\mathbf{w}_k}(x)] )$;
${\rm Var}\{f_{\mathbf{w}_k}|x\} =  \mathbb{E}_{ \mathbf{{w}_k} } [(f_{\mathbf{w}_k}(x) - \mathbb{E}_{ \mathbf{{w}_k} }[f_{\mathbf{w}_k}(x)])^2]$;
${\rm Cov}\{f_{\mathbf{w}_k,\mathbf{w}_k^\prime}|x\} =  \mathbb{E}_{ \mathbf{{w}_k},\mathbf{{w}_{k^\prime}}}
    [ (f_{\mathbf{w}_k}(x) - \mathbb{E}_{ \mathbf{{w}_k} }[f_{\mathbf{w}_k}(x)])(f_{\mathbf{w}_{k^\prime}}(x) - \mathbb{E}_{ \mathbf{{w}_{k^\prime}} }[f_{\mathbf{w}_{k^\prime}}(x)])]$.  
\label{theorem1}
\end{theorem}

In Theorem \ref{theorem1}, the underlying  meanings of the five factors are elaborated as follows:
\begin{itemize}
    \item  ${\rm{TrainBias}} \{f_{\mathbf{w}_k}|(x,y)\} $  measures the   fitting capability of  a client model ${\mathbf{w}_k}$ on  the samples of client dataset  (i.e.,  $(x,y) \in \mathcal{D}_k$), where  ${\mathbf{w}_k}$ is trained on $\mathcal{D}_k$;
    \item  ${\rm{HeterBias}} \{f_{\mathbf{w}_k}|(x,y)\}$ measures the degree of catastrophic forgetting of a client model ${\mathbf{w}_k}$ on the non-overlapping samples between the global dataset and client dataset (i.e., $(x,y)\in\mathcal{D}\setminus\mathcal{D}_k$), where ${\mathbf{w}_k}$ is trained on $\mathcal{D}_k$;
    \item ${\rm Var}\{f_{\mathbf{w}_k}|x\}$  measures the sensitivity of a client model $\mathbf{w}_k$ to small fluctuations in the given sample input   $ x \in \mathcal{D}_k$, which does not depend  on   $y$;
    \item  ${\rm Cov}\{f_{\mathbf{w}_k,\mathbf{w}_k^\prime}|x\}$ denotes the output correlation  between client models $ \mathbf{w}_k $ and $ \mathbf{w}_k^\prime$ given the same input $x$, which does not depend  on $y$;
    \item  $O(\Delta^2 )$ represents the locality in \cite{izmailov2018averaging,rame2022diverse}, i.e., the maximum distance between client and global models.
\end{itemize}
Based on these five factors, the capability of the global model on the global dataset can be quantified by the capabilities of client models on their local datasets.
Due to the presence of unseen samples for the client model $\mathbf{w}_k$ in its local dataset, ${\rm{HeterBias}} \{f_{\mathbf{w}_k}|(x,y)\}$ is expected to have a more significant impact on the global model compared to ${\rm{TrainBias}} \{f_{\mathbf{w}_k}|(x,y)\}$.
This implies that client models that are more robust to catastrophic forgetting contribute to a lower loss for their corresponding global model.
In other words, HeterBias can effectively measure the performance of the global model on the client side.
Moreover, unlike the decomposition in (\ref{decomposistion:step1}), Theorem \ref{theorem1} incorporates a covariance term to account for the fact that client models are not independent and identically distributed within a model solution subspace due to the heterogeneity of the data.
In the following, we empirically validate the effect of these factors on global model capability throughout the training.

 \subsection{Empirical Validation of Decomposition Analysis}
The FL setup involves training a global model on CIFAR-10 across ten clients for 400 rounds, where clients hold two class shards of CIFAR-10 for heterogeneous data setup (see specific setups in Table  \ref{Setup_all_visuz_exps}).

\subsubsection{The effect of bias factor: heterogeneous bias dominates the loss of the global model after the early training}

\begin{figure}[t]
 \begin{center}
		\includegraphics[width=0.3\textwidth]{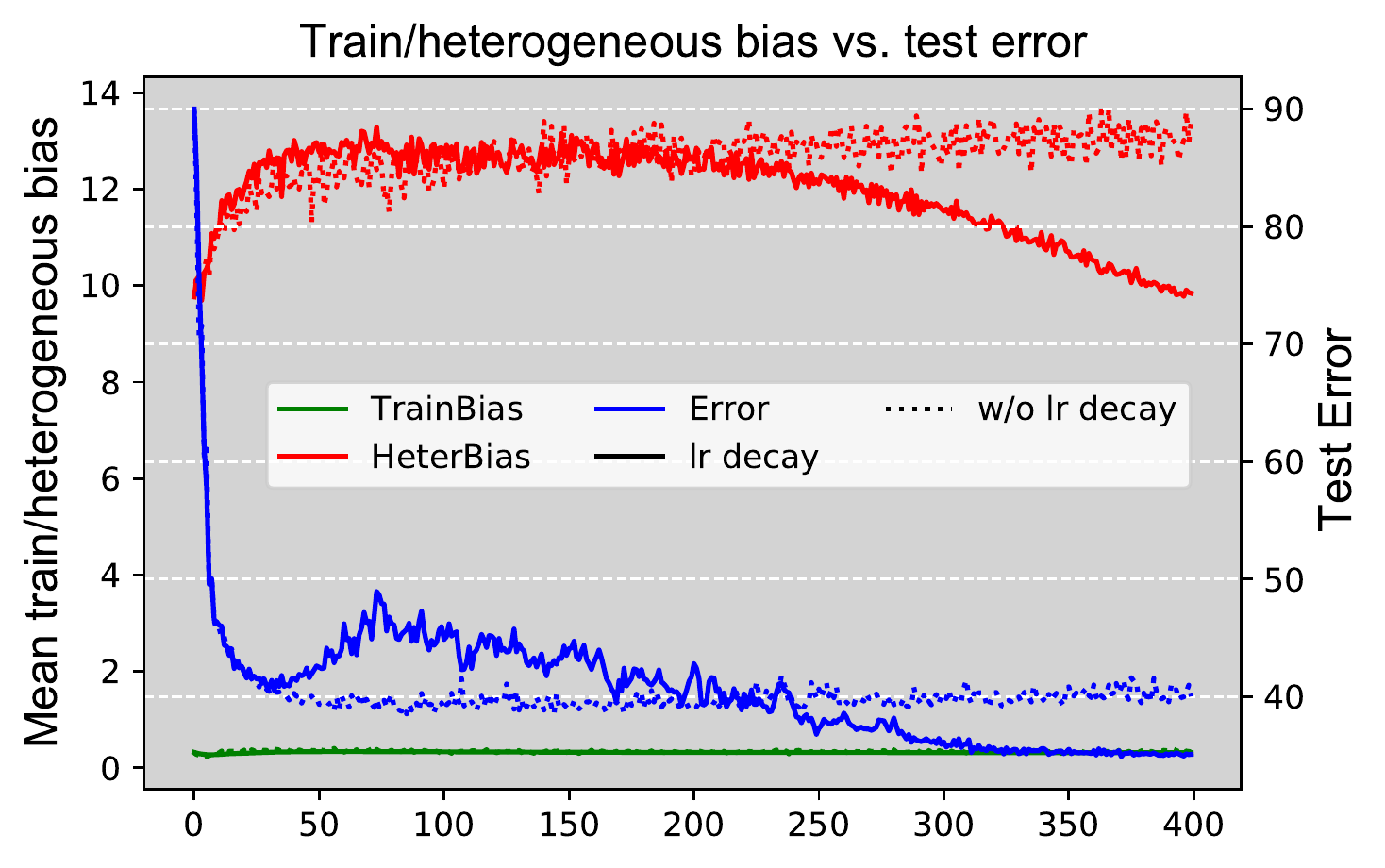}
  \vspace{-5pt}
		\caption{Train and heterogeneous biases  w.r.t rounds (\textit{x-axis}).}
  		\label{Bias_validation} 
\end{center}
\vspace{-10pt}
\end{figure}

 Figure \ref{Bias_validation} shows that the   ${\rm{TrainBias}} \{f_{\mathbf{w}_k}|(x,y)\} $ is effectively reduced to almost zero, which is because the number of local updates is sufficient for client models to fit their datasets.
However, heterogeneous data introduce a non-zero ${\rm{HeterBias}} \{f_{\mathbf{w}_k}|(x,y)\}$, which individual client struggles to address through local training due to missing samples from the global dataset.
Nonetheless, due to the geometric property observed in  Section \ref{section:loss_landscape}, FMA provides an initialization point with enriched global information for client models to mitigate this bias.   

Moreover,  the larger the local update step, the more global information the FMA provides is forgotten, and the greater the heterogeneous bias becomes due to the catastrophic forgetting phenomenon in neural networks \cite{goodfellow2013empirical, kirkpatrick2017overcoming}.
This is validated by the cases with and without learning rate (lr) decay shown in Figure \ref{Bias_validation}.
We use a round-exponential decay lr to control the update size, a straightforward approach to preventing catastrophic forgetting in FL  \cite{reddi2021adaptive}. 
In the early phase, heterogeneous bias does not significantly impact the test classification error because the error continues to decrease even if the bias increases. 
However, both the error and the bias show a positive correlation in both cases.
For example,  after approximately 40 rounds, they both increase and decrease in the case of lr decay, and the error grows slightly with the bias in the case without lr decay.

\subsubsection{The effect of locality factor: controlling the locality helps reduce the global model loss at the late training}   
\begin{figure}[t]
 \begin{center}
		\includegraphics[width=0.3\textwidth]{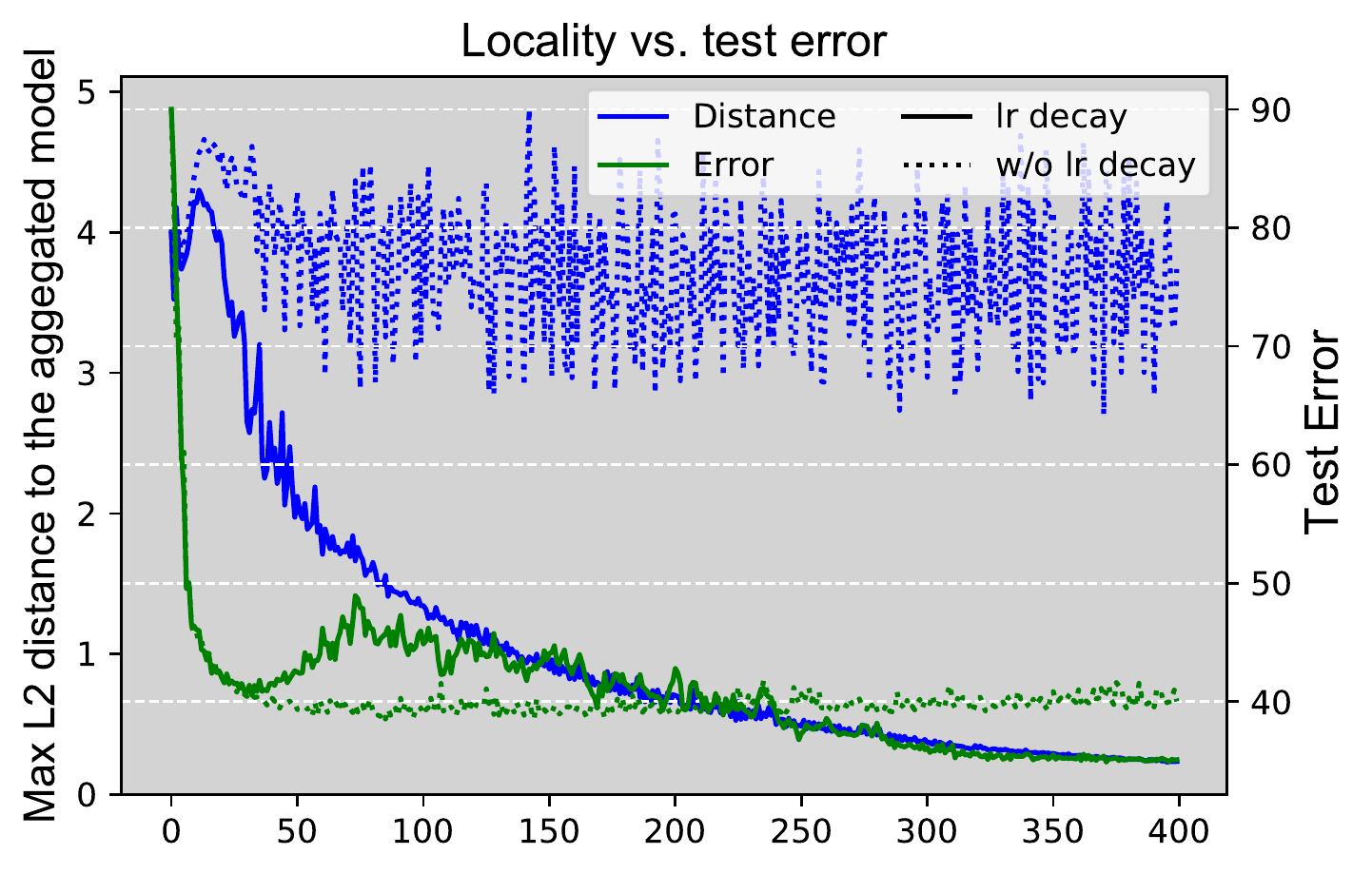}
    \vspace{-5pt}
		\caption{Locality (\textit{L2 distance}) w.r.t rounds (\textit{x-axis}).}
  		\label{locality_validation} 
\end{center}
\end{figure}

In Figure \ref{locality_validation}, we employ the L2 distance to quantify the locality term $O(\Delta^2)$.
 Theorem \ref{theorem1} demonstrates that the test loss decreases as the maximum distance between client models and the global model, i.e., $\Delta$, reduces.

 Figure \ref{locality_validation}
  shows that the locality is larger in the case without lr decay, which results in a more significant test error. 
Before the 40th round, the test classification errors of both cases continue to decline despite an increase in the locality occurring within this period. 
Then,   in the case of lr decay, the locality reduces while the error increases from 40 to 75 rounds.
This indicates that the locality does not correlate strongly with the error during the early training.
The locality stabilizes after the early training phase (i.e., the locality is upper-bounded in both cases).
This further validates the proximity of client models to the global model, as discussed in Section \ref{section:loss_landscape}.


\subsubsection{The effect of variance factor: reducing global model loss by aggregating more client models in FMA is limited} 
When the client dataset $\mathcal{D}_k$ does not change during the training, the variance factor $\mathbb{E}_{(x,y)\in \mathcal{D}}[{\rm Var}\{f_{\mathbf{w}_k}|x\}]= 1/n_k\sum_{(x,y)\in\mathcal{D}_k} {\rm Var}\{f_{\mathbf{w}_k}|x\} $ in Theorem \ref{theorem1} can be viewed as a constant $V_k$ since  ${\rm Var}\{f_{\mathbf{w}_k}|x\}$  depends on the discrepancy between $\mathcal{D}$ and $\mathcal{D}_k$.
Specifically,  from  Theorem 1 in \cite{ye2022ood} and Proposition 2 in \cite{rame2022diverse}, we have the following property:
\begin{theorem}\label{Theorem:Bounded_variance} (Bounded variance.)
Given a kernel regime $f_{\mathbf{w}_k}$ trained on client dataset $\mathcal{D}_k$ (of size $n_k$) with  neural tangent kernel $K_{f_{\mathbf{w}_k}}$, when $\exists ( \lambda_{\mathcal{D}_k},\epsilon) $ with $0\leq\epsilon \ll \lambda_{\mathcal{D}_k}$ such that $\forall x_i \in \mathcal{D}_k$, $K_{f_{\mathbf{w}_k}}(x_i,x_i)=\lambda_{\mathcal{D}_k}$ and $\forall x_i, x_j \in \mathcal{D}_k$ and $i\neq j$, $|K_f(x_i,x_j) | \leq \epsilon$, the variance on the global dataset $\mathcal{D}$ is:
\begin{equation}
\begin{aligned}
\mathbb{E}_{x \in \mathcal{D}}[{\rm Var}\{f_{\mathbf{w}_k}|x\}]
= &\frac{n_k}{2\lambda_{\mathcal{D}_k}} MMD^2(\mathcal{D}_k,\mathcal{D} )
\\& + \lambda_{\mathcal{D}} 
- \frac{n_k}{2\lambda_{\mathcal{D}_k}}\beta_\mathcal{D}  + O(\epsilon),
\end{aligned}
\end{equation}
where $MMD(\cdot)$ is the empirical maximum mean discrepancy in the reproducing kernel Hilbert space (RKHS) of $K_{f_{\mathbf{w}_k}}(x_i,x_j)$; $\lambda_{\mathcal{D}} = \mathbb{E}_{x \in \mathcal{D}}K_{f_{\mathbf{w}_k}}(x,x)$ and $\beta_\mathcal{D}=\mathbb{E}_{x_i,x_j \in \mathcal{D}, i\neq j}K^2_{f_{\mathbf{w}_k}}(x_i,x_j)$ denote the empirical mean similarities of identical and different samples averaged over $\mathcal{D}$, respectively.
\label{theorem3}
\end{theorem}
In   Theorem \ref{theorem3},   both $\lambda_{\mathcal{D}}$ and $\beta_\mathcal{D}$ depend exclusively on the global dataset $\mathcal{D}$ for a  $f_{\mathbf{w}_k}$.
The global dataset $\mathcal{D}$ represents the combination of all client datasets and can be viewed as a fixed dataset in FL.
Consequently,  $\lambda_{\mathcal{D}}$ and $\beta_\mathcal{D}$ can be regarded as   constants that depend  on $\mathcal{D}$ in Theorem \ref{theorem3}.
Therefore, Theorem \ref{theorem3} demonstrates that the variance term in Theorem \ref{theorem1} is solely associated with $MMD^2(\mathcal{D}_k,\mathcal{D})$, which quantifies the distance between the client dataset $\mathcal{D}_k$ and the global dataset $\mathcal{D}$ in FL setups.

\begin{figure}[t]
 \begin{center}
		\includegraphics[width=0.5\textwidth]{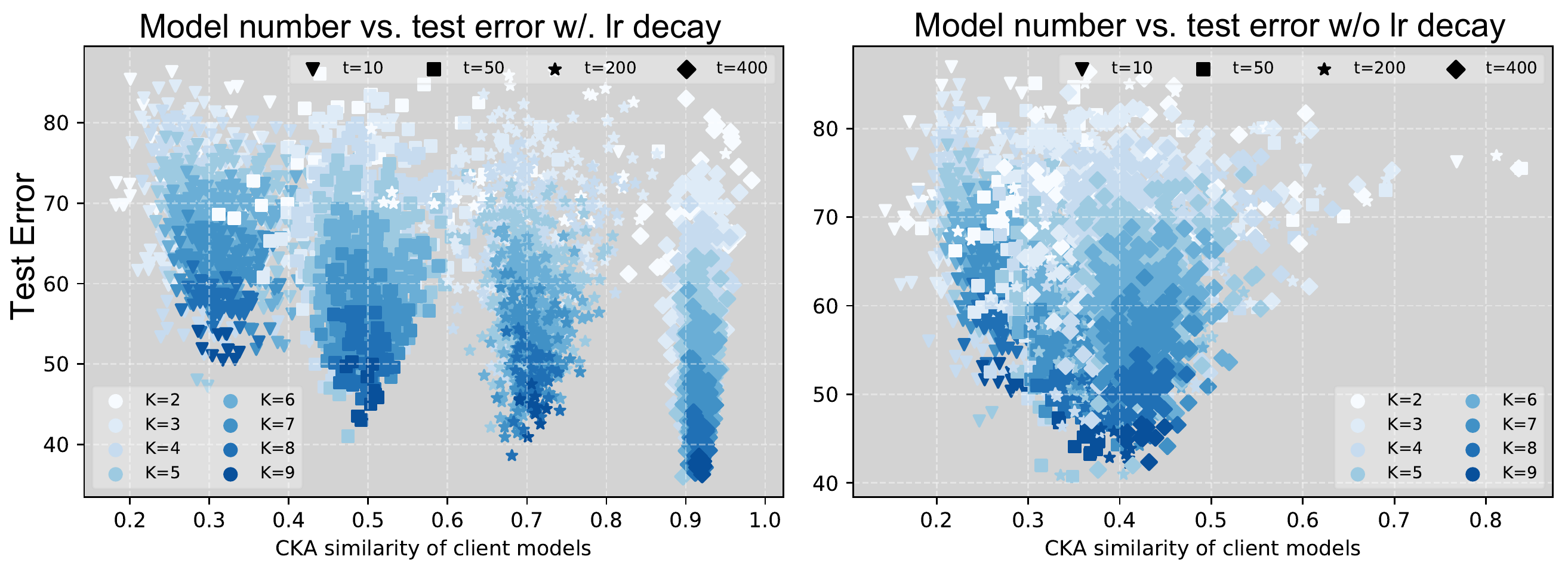}
    \caption{Test error w.r.t model number and model similarity.}
  \label{Variance_covariance_validation} 
\end{center}
\end{figure}

From  Theorem \ref{theorem3}, we have $ \max_k\{V_k\} \leq \epsilon$. 
The whole variance term in Theorem \ref{theorem1} is $ \sum_k{n_k^2}/{n^2}V_k$ and it is upper bounded by   $ \epsilon\sum_{k=1}^{K}{n_k^2}/{n^2}$.
Then, FMA can keep this upper bound diminishing by averaging more client models  (i.e., larger $K$ induces smaller $\sum_{k=1}^{K}{n_k^2}/{n^2}$)  to reduce the global model loss when $\epsilon$ is tight for  $\max_k\{V_k\}$.
Figure  \ref{Variance_covariance_validation} verifies this effect on FMA throughout the training process.
However, it is important to note that the impact of the client number becomes negligible when all client models are identically distributed (i.e., client models are trained on homogeneous datasets with the same training configurations). 
This is because the sum of the third and fourth terms on the right-hand side of Theorem \ref{theorem1} (i.e., ${\rm Var}\{f_{ \mathbf{w}_k}|x\}= {\rm Cov}\{f_{\mathbf{w}_k,\mathbf{w}_{k^\prime}}|x\}$ when client models are identically distributed) is equal to the variance of a single client model.

In summary, the variance term decreases as the number of client models being averaged in FMA increases. 
Nonetheless, this effect weakens as more models are incorporated.

\subsubsection{The effect of covariance factor: heterogeneous data inherently results in a small but lower bounded covariance}

To measure the covariance factor ${\rm Cov}\{f_{\mathbf{w}_k,\mathbf{w}_k^\prime}|x\}$,  we employ the CKA similarity \cite{krizhevsky2009learning} to compute the output correlation among client models given the same input. 
As shown in Figure \ref{Variance_covariance_validation}, heterogeneous data inherently lead to a small covariance term, especially for the case without lr decay.
 That is,  maintaining high diversity among client models (e.g., \cite{lee2016stochastic,rame2022diverse}) may not significantly reduce the loss of the global model. 
 Indeed,  it can negatively impact the performance in the late training stage, as illustrated by the comparison between both cases at the 400th round in Figure \ref{Variance_covariance_validation}.

Furthermore, we show that the covariance term has a non-zero lower bound that depends on the maximum discrepancy across client datasets.
Let $n_i = n_j$, $\forall i, j \in [K]$ (i.e., the number of client samples is the same). By ablating the impact of weighted averaging, we can further decompose the covariance term in Theorem \ref{theorem1} as follows: 
\begin{corollary} (Lower bound of the covariance term.) For $n_i = n_j$ when $i\neq j$, the covariance term in Theorem \ref{theorem1} is bounded by:
    \begin{equation}
    \begin{aligned}
\mathbb{E}_{(x,y) \in \mathcal{D}} & ( \frac{1}{K^2} {\sum_k \sum_{{k^\prime}}  {\rm Cov}\{f_{\mathbf{w}_k,\mathbf{w}_{k^\prime}}|x\}}) 
\\ &= \frac{1}{nK^2}\sum_{(x,y)\in\mathcal{D}} {\sum_k \sum_{{k^\prime}}    {\rm Cov}\{f_{\mathbf{w}_k,\mathbf{w}_{k^\prime}}|x\}}
\\& \geq \frac{K-1}{nK}\sum_{(x,y)\in\mathcal{D}} \min_{(k,{k^\prime})}   {\rm Cov}\{f_{\mathbf{w}_k,\mathbf{w}_{k^\prime}}|x\},
    \end{aligned}
\end{equation}
where $\min_{(k,{k^\prime})}   {\rm Cov}\{f_{\mathbf{w}_k,\mathbf{w}_{k^\prime}}|x\}$ measures the maximum discrepancy  among all client models. 
  \label{further_covariance}
\end{corollary}
The physical meaning of $\min_{(k,{k^\prime})}   {\rm Cov}\{f_{\mathbf{w}_k,\mathbf{w}_{k^\prime}}|x\}$, when given a sample $(x,y) \in \mathcal{D}$, can be understood as follows:
Firstly,   ${\rm Cov}\{f_{\mathbf{w}_k,\mathbf{w}_{k^\prime}}|x\}$ calculates the covariance between   client models $\mathbf{w}_k$ and $\mathbf{w}_{k^\prime}$.
Then, $\min_{(k,{k^\prime})}   {\rm Cov}\{f_{\mathbf{w}_k,\mathbf{w}_{k^\prime}}|x\}$ finds the minimal value of  ${\rm Cov}\{f_{\mathbf{w}_k,\mathbf{w}_{k^\prime}}|x\}$ across all client pairs $(k,k^\prime) $, where $  \forall k, k^\prime \in [K],  k \neq k^\prime$.
This minimal value measures the largest diversity among client models on the given sample $(x,y)$.
The maximum discrepancy across client datasets determines the diversity and remains constant since client datasets do not change in the generic FL setups.

Therefore,   Corollary \ref{further_covariance}  demonstrates that the covariance term has a lower bound that depends on the maximum discrepancy across client datasets.
Consequently, the effect of FMA on reducing the loss of the global model by controlling the diversity of client models is limited.

\subsubsection{Summary}\label{sec:summary}
From the above discussion, we summarize the impact of the five factors in Theorem \ref{theorem1} on the loss of the global model during training as follows:
\begin{itemize}
    \item ${\rm{TrainBias}} \{f_{\mathbf{w}_k}|(x,y)\} $  keeps almost zero throughout the training process;
    \item  ${\rm{HeterBias}}$ $\{f_{\mathbf{w}_k}|(x,y)\}$   and  $O(\Delta^2 )$ dominate the loss of the global model after the early training; 
    \item  The weighted sum of ${\rm Var}\{f_{\mathbf{w}_k}|x\}$ can be reduced to some extent with a large number of client models;
    \item ${\rm Cov}\{f_{\mathbf{w}_k,\mathbf{w}_k^\prime}|x\}$ is too small to affect the loss of the global model.
\end{itemize}
Therefore,   FMA  can reduce the loss of the global model by controlling ${\rm{HeterBias}}$ $\{f_{\mathbf{w}_k}|(x,y)\}$  and locality $O(\Delta^2 )$, in addition to aggregating more client models.

\section{Proposed Method}\label{section:proposed_method}
In this section, we will begin by discussing our motivation, i.e., to alleviate the deviation of global models in FMA. After that, we will introduce IMA and mild client exploration to address this deviation. Lastly, we will discuss the advantages of using IMA for FL.

\subsection{Motivation: Deviation of Global Models from the Basin's Center} \label{section:motivation_IMA}

\begin{figure*}[t]
	\centering  
	\subfigbottomskip=2pt 
	\subfigcapskip=-5pt 
 	\subfigure[IMA's Toy example]{
		\includegraphics[width=0.23\linewidth]{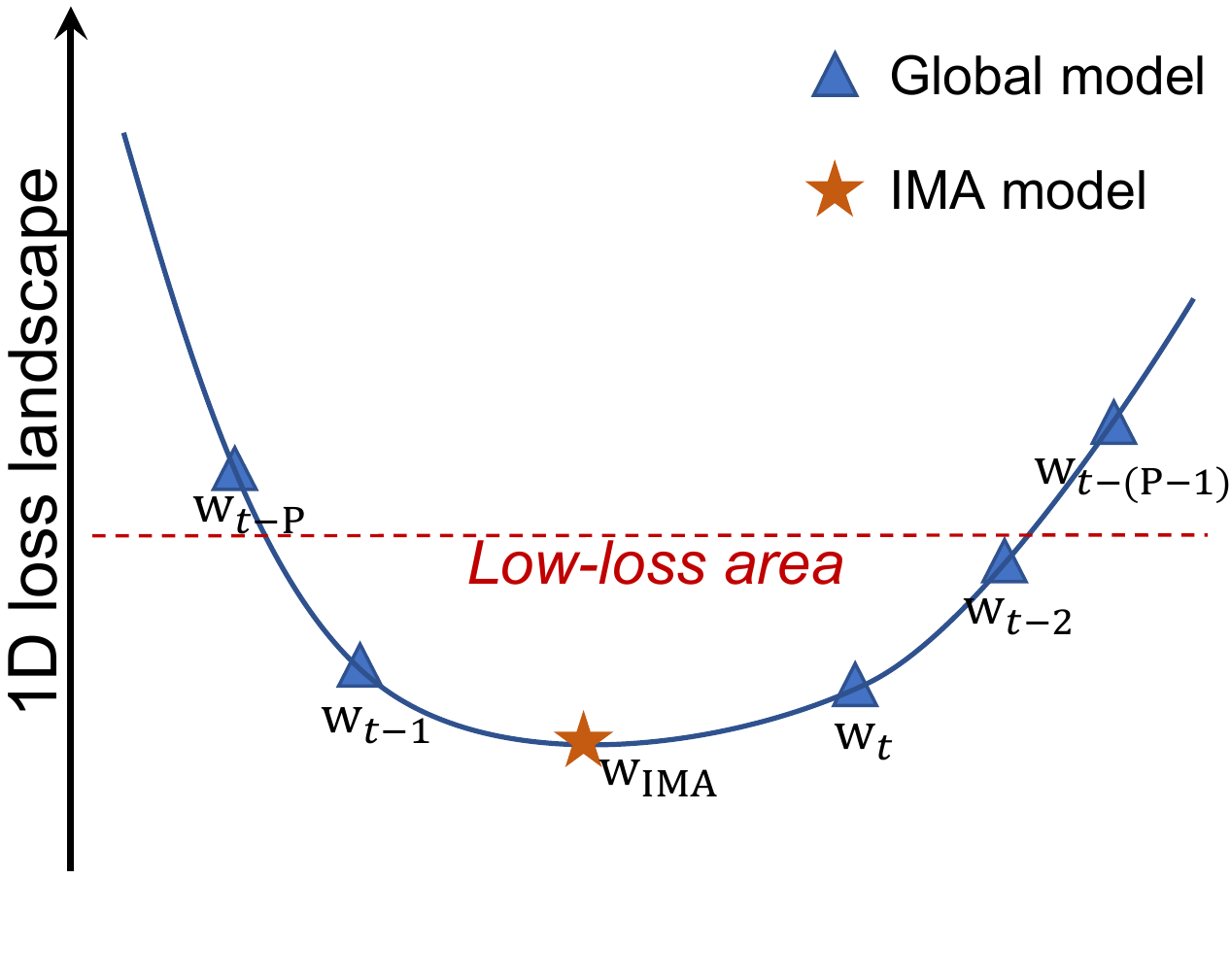}
  \label{IMA_validation_toy_example}}
	\subfigure[Interpolation among global models]{
		\includegraphics[width=0.36\linewidth]{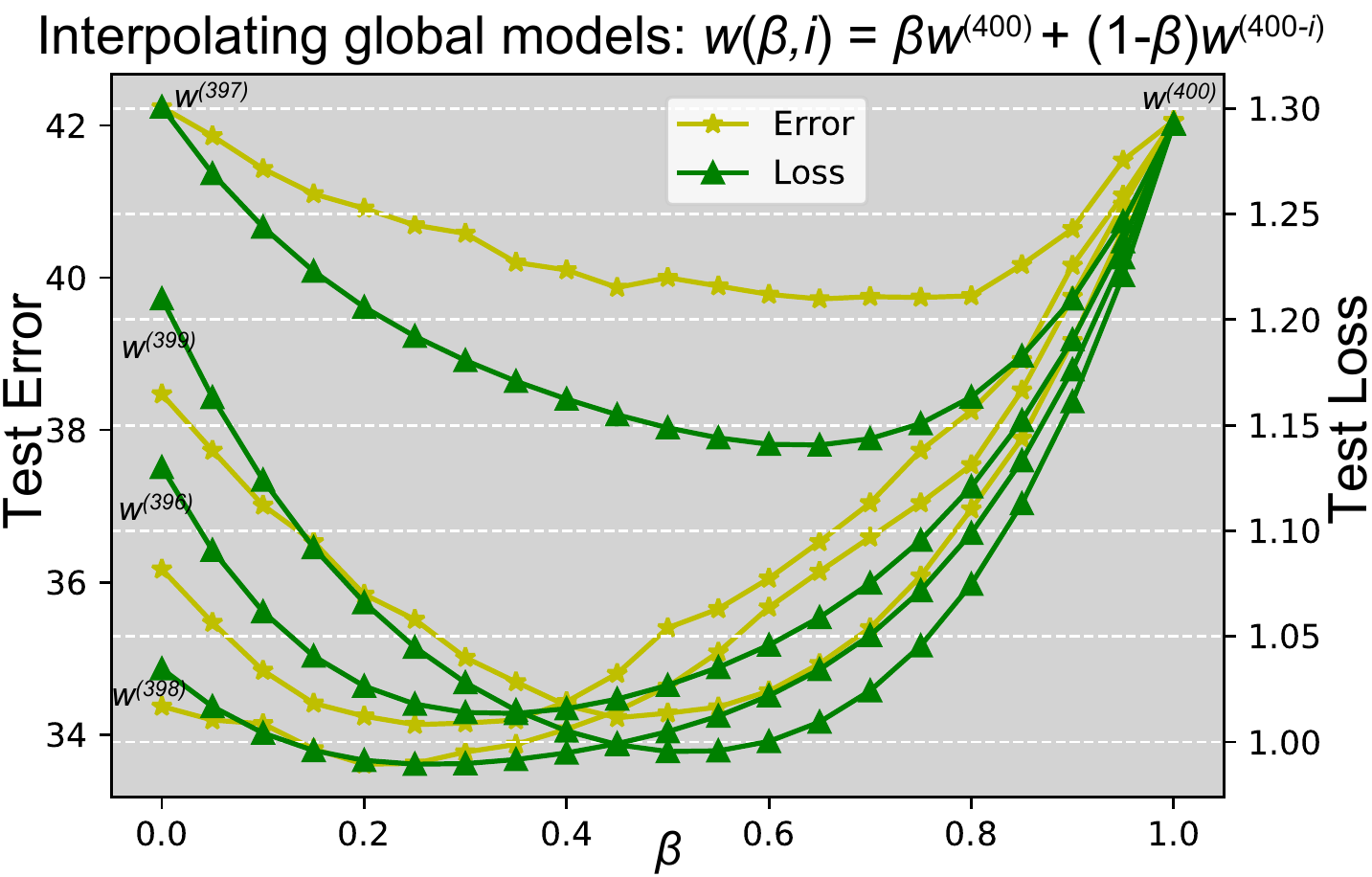}
    \label{IMA_validation_interpolation}}
	\subfigure[Interpolation global and IMA models]{
		\includegraphics[width=0.36\linewidth]{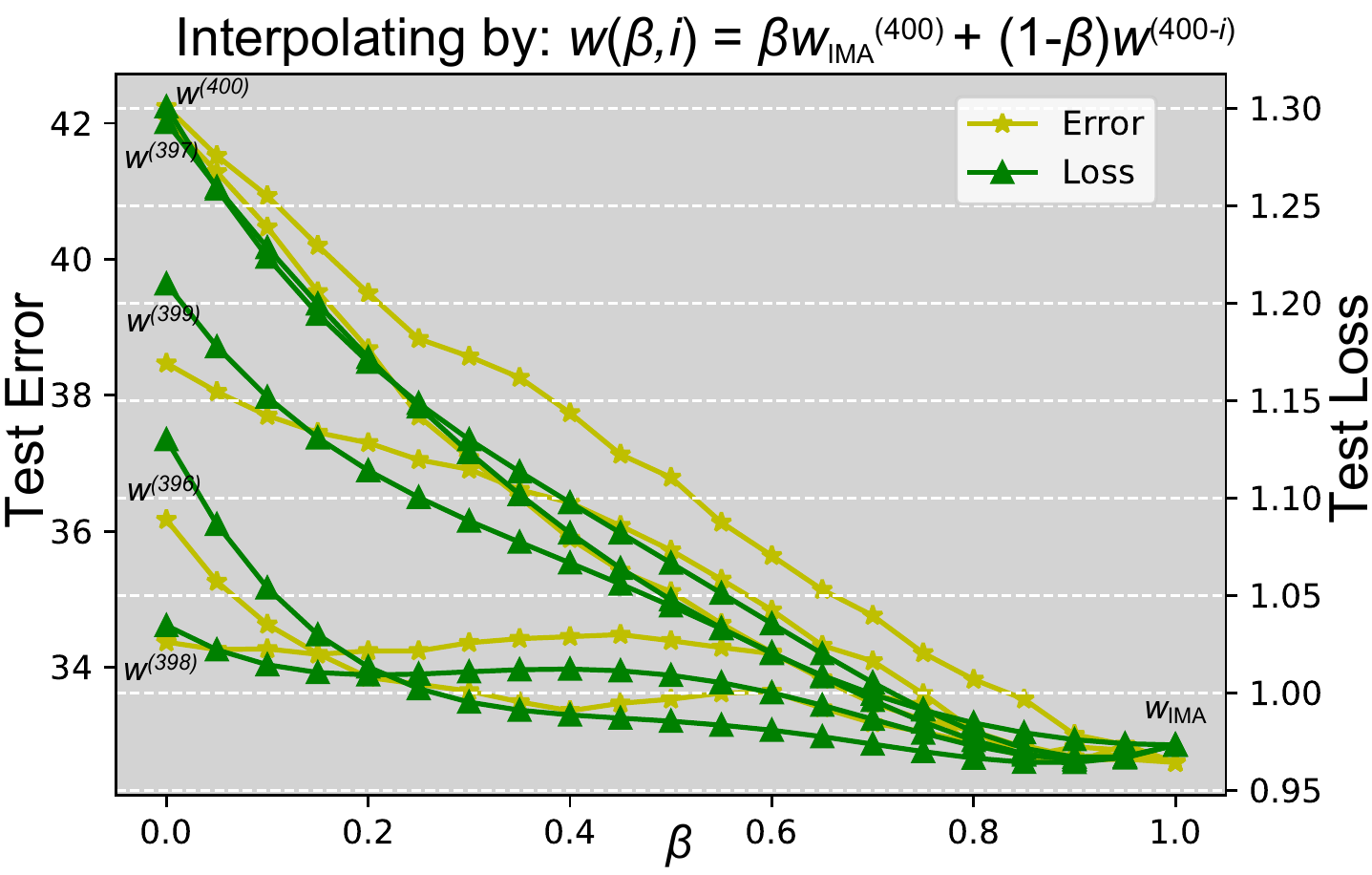}
    \label{Loss_flatness_interpolation}}
  \caption{(a) A toy example of 1D loss landscape visualization to show the motivation of IMA; (b) interpolation among global models to validate the effect of interpolation on alleviating global models' deviation from the basin's center; (c) interpolation between the IMA  and global models to indicate the flatness of the IMA model near the basin's center.}
  \label{IMA_validation}
\end{figure*}

As shown in the fourth column of Figure \ref{loss_error_visualization}, when performing FMA on heterogeneous data, global models deviate from the basin center of the loss landscape.
Specifically, FMA tends to move the global model towards the center of the loss basin.
However, due to heterogeneous data, the movement of the global model deviates from the basin's center. This deviation causes global models obtained in different rounds to be scattered around the basin's center, as illustrated in Figure \ref{IMA_validation_toy_example}.
This geometric property can be leveraged to improve the aggregation process and bring the global model closer to the basin's center.

The loss decomposition analysis of the global model presented in Theorem \ref{theorem1} provides a new perspective on this geometric property.
In FL, a small number of clients participating in each round makes the one-cohort dataset of participating clients $ \mathcal{D}_C^{(t)} = \cup_{i=1}^C\mathcal{D}_i$  inconsistent from the global dataset $\mathcal{D}$. 
Moreover, the weighted averaging tends to assign higher weights to clients with datasets that are large but imbalanced compared to  $\mathcal{D}$, further exacerbating the inconsistency between $\mathcal{D}_C^{(t)} $ and $\mathcal{D}$.
Consequently,  heterogeneous bias cannot be completely reduced since the one-cohort dataset misses data samples  $(x,y)\in\mathcal{D}^{(t)} \setminus\mathcal{D}_C$. 
This leads to the observed deviation of global models from the basin's center.

In contrast, a combination of one-cohort datasets from different rounds, denoted by $\mathcal{D}_{\rm IMA} = \cup_{i=0}^{P-1}\mathcal{D}_C^{(t-i)}$, contains fewer missing data samples than $\mathcal{D}^{(t)}$ alone.
As summarized in Section \ref{sec:summary}, reducing heterogeneous bias can decrease the global model's loss.
This implies that aggregating historical global models can reduce the heterogeneous bias on missing data samples since the global model $\mathbf{w}^{(t-i)}$ carries the information of $\mathcal{D}_C^{(t-i)}$.
To verify this, we linearly interpolate two global models from different rounds and evaluate the performance of the interpolated models on the CIFAR-10 dataset, as depicted in Figure \ref{IMA_validation_interpolation}.
The figure demonstrates that lower loss/error points consistently exist within the global models'  interpolation. 
In other words, interpolated models retain more global information than a solo global model while remaining within a common basin, thus reducing the heterogeneous bias.
Therefore, we apply Theorem \ref{theorem1} to leverage the geometric properties of FMA to alleviate the deviation of global models.

 \subsection{Iterative Moving Averaging (IMA)} 
The missing information of one-cohort client models on $(x,y)\in\mathcal{D}\setminus\mathcal{D}_C$ can be compensated by utilizing historical global models.
This compensation can be achieved by aggregating historical global models into the latest one, as supported by the observation in Figure \ref{IMA_validation_interpolation}.
Therefore, we propose applying IMA to historical global models after sufficient training rounds instead of ignoring them in conventional  FMA.
Specifically, as illustrated in Figure \ref{IMA_validation_toy_example}, after  $t_s$ rounds, the server performs FMA with IMA to obtain an averaged model from a time window of previous rounds as:
\begin{equation}
    \mathbf{w}_{{\rm IMA}}^{(t)} \leftarrow \frac{1}{P} \sum_{i=0}^{P-1} \mathbf{w}^{(t-i)}, \quad \quad t \geq t_s,
    \label{iterative model averaging}
\end{equation}
where $\mathbf{w}_{{\rm IMA}}^{(t)}$ is the IMA model for the $t$-th round,   $P$ is the size of the time window, and $t_s$ is the starting round of IMA.
The complete process of  IMA is illustrated in Algorithm \ref{alg:IMA}.

To mitigate the impact of information noise introduced by historical global models, we initiate IMA in the later training phase, such as $0.75R$ with $R$ denoting the total number of training rounds.
Note that IMA provides a better initialization for client models to perform local training, thus resulting in faster convergence and higher accuracy.
Importantly,  IMA only requires storing $P$ global models $\{\mathbf{w}^{(\tau)}\}_{\tau =t-P}^t$  obtained by FMA and initializing client models with $\mathbf{w}_{{\rm IMA}}^{(t)}$ for the next round, without modifying client participation or weighted aggregation. 
Consequently,   IMA can be readily integrated into various FL methods to maintain the global model within the low-loss landscape region, as demonstrated in Figures \ref{Loss_flatness_interpolation} and \ref{fig:loss_landscape_two_model}.

\subsection{Mild Client Exploration in IMA}
Theorem \ref{theorem1} indicates that controlling the locality can reduce the loss of the global model in the late training stage.
Based on this insight and the geometric properties discussed in Section \ref{section:loss_landscape}, we highlight the importance of regulating the magnitude of client updates once the global model enters the low-loss area after sufficient training rounds, as illustrated in Figure \ref{IMA_validation_toy_example}.
Otherwise, clients may converge to their local optimal models such that the global model deviates from the low-loss area. 
This is because these local models reach the wall of the loss basin of the global model instead of the bottom, as suggested by the geometric properties of FMA, even when they are close to the global model.

To address this issue, we adopt a more aggressive learning rate decay, called mild client exploration, to control updates during late training.
This involves a significant exponential lr decay, such as 0.03 lr decay per round. 
  Table \ref{table:decaying_schemes} demonstrates that some methods, using a small and constant lr in IMA, yield similar results to ours when they sufficiently constrain client updates. 
In contrast, as shown in Table \ref{table:decaying_schemes}, when the locality is not adequately controlled during late training (i.e., non-additional lr decay),    the deviation of the global model in FMA may impact its performance.

\subsection{Advantages of IMA for FL}
In contrast to FMA without considering previous global models,  IMA  uses a sliding window to average the global models over successive training rounds.
As discussed in Section \ref{section:motivation_IMA}, FMA may deviate the global model from the expected loss-basin center when facing heterogeneous data.
Therefore, IMA  is built upon FMA and leverages the geometric property of global models in the loss landscape to mitigate the deviation introduced by FMA, as validated in Figures \ref{IMA_validation_interpolation} and \ref{Loss_flatness_interpolation}.

SWA \cite{izmailov2018averaging} has been widely used to enhance the model performance by aggregating the training checkpoints when convergence is near. Compared with SWA, IMA collects $P$ previous global models when the FL training has not converged. In addition, IMA and SWA apply to different scenarios: SWA involves centralized training on homogeneous data, while IMA is tailored for FL on heterogeneous data.  
Specifically, due to heterogeneous data,  FMA causes the global model to deviate towards the wall of the loss basin, instead of the expected basin center, as illustrated in Figure  \ref{IMA_validation_toy_example}.
Notably, the deviation causes the global models to surround the center from different rounds. 
IMA leverages this geometric property to average global models over a time window,  bringing the IMA model closer to the loss-basin center while avoiding the injection of outdated information.
Then,  IMA re-initiates the client models with the IMA model, whereas SWA does not.
The re-initiation with the IMA model corrects the training trajectory and speeds up FL training.

Moreover, it is worth noting that model compression techniques, such as sparsification \cite{tenison2022gradient,PandaMBCM22}, quantization \cite{ShlezingerCEPC21}, and low-rank decomposition \cite{lan2023communication}, can be seamlessly integrated with IMA to further reduce the communication overhead and accelerate FL training on heterogeneous data.
These techniques operate orthogonally to IMA, which focuses on correcting the trajectory of the global model in the loss landscape.
By combining IMA with model compression, we can achieve a two-fold benefit: mitigating the impact of heterogeneous data on model convergence and reducing the communication burden. This synergistic effect is exemplified in Table \ref{table:Communication_effeciency}, which demonstrates that IMA is compatible with FedGMA, a method that employs AND-Masked sparsification to accelerate the FL training.

 \begin{algorithm}[t]
   \caption{FL with IMA}
   \label{alg:IMA}
\begin{algorithmic}
   \STATE {\bfseries  Input:} model $\mathbf{w}$,  total client number \textit{K}, IMA's start round  $t_s$, IMA's window size $P$
   \FOR{ each round $t=1,\cdots, R$  }
 
\STATE Server    samples clients $\mathcal{S} \subseteq [K]$
    \STATE \textbf{if} $t \geq t_s$ \textbf{do}
     \STATE  \quad  Server   sends  $\mathbf{w}_{\rm IMA}^{(t-1)}$ to all clients $i\in \mathcal{S}$
    \STATE \textbf{else}:  
   \STATE  \quad  Server   sends  $\mathbf{w}^{(t-1)}$ to all clients $i\in \mathcal{S}$
   \STATE \textbf{on client }$i\in \mathcal{S}$ \textbf{in parallel do}
   \STATE \quad  \textbf{if} $t\geq t_s$ \textbf{do}
      \STATE \quad \quad Initialize the local model  $\mathbf{w}_i \gets \mathbf{w}_{\rm IMA}^{(t-1)}$ 
    \STATE \quad \quad Local training with mild exploration and get $\mathbf{w}_i^{(t)}$
    \STATE  \quad \textbf{else}: 
        \STATE \quad \quad Initialize the local model $\mathbf{w}_i \gets \mathbf{w}^{(t-1)}$  
      \STATE  \quad  \quad Local training  and get $\mathbf{w}_i^{(t)}$
    
    \STATE \quad  \textbf{end for}
    \STATE \quad  Send $\mathbf{w}_i^{(t)} $  back to the server
    \STATE   \textbf{end on client}
    \STATE Server performs  FMA $\mathbf{w}^{(t)} \gets   \sum_{i\in \mathcal{S}} (n_i/ \sum_{i\in\mathcal{S}}n_i)\mathbf{w}_i^{(t)}$ 
    \STATE  \textbf{if} $t\geq t_s$ \textbf{do}
     \STATE \quad   Server performs IMA  $ \mathbf{w}_{{\rm IMA}}^{(t)} \leftarrow \frac{1}{P} \sum_{\tau=0}^{P-1} \mathbf{w}^{(t-\tau)}$
   \ENDFOR
\end{algorithmic}
\end{algorithm}

\section{Experiments}\label{section:Experiments}
In this section, we present experimental results to verify the effectiveness of IMA by comparing it with existing methods.
We will first describe the experimental setups. Next, we will present results on different heterogeneous data setups, datasets, models,  FL setups, and baselines. 
Finally, we conduct a comprehensive ablation study on IMA, including different starting rounds, window sizes, and lr decays.

\subsection{Experimental Setups}
\subsubsection{Heterogeneous Data Setups}
We examine label/feature distribution skew in heterogeneous data \cite{kairouz2021advances} and refer to them as label/feature skew.
To simulate label skew,  we divide  FMNIST \cite{xiao2017fashion} and CIFAR-10/100 into \textit{data shards with the same sample number} for clients  (e.g., $\#C=2$ indicates that each client holds two classes as in \cite{mcmahan2017communication}). 
We use the Dirichlet distribution $Dir(\alpha)$ to create client datasets with \textit{different sample numbers} according to \cite{yurochkin2019bayesian}. 
Moreover, we combine label skew and feature skew on Digit Fives \cite{li2021fedbn} and PACS \cite{li2017deeper}.
Specifically, we divide each domain dataset (i.e., feature skew)  into 20 subsets, each for one client, with diverse label distributions  (i.e., label skew).
The combined skew on Digit Fives and PACS is a more heterogeneous case than their inherent feature domain shift.

\begin{table*}[t]
\centering
\caption{Mean top-1 last-10-round accuracy comparison of all methods with and without IMA under label skew (including $\#C=2$ and $\alpha=0.1$) and feature skew (FS). We follow \cite{reddi2021adaptive} to use Pachinko Allocation (PA) \cite{li2006pachinko} to create a federated CIFAR-100. 
Bold text indicates the best results between IMA and IMA-free methods, while underlined text denotes the best results with or without IMA.}
\label{table:performance}
\resizebox{\textwidth}{!}{%
\begin{tabular}{@{}cccccccccc@{}}
\toprule
\begin{tabular}[c]{@{}c@{}}Dataset\\ (Model)\end{tabular} & \begin{tabular}[c]{@{}c@{}}Heter\\ Data\end{tabular} & \begin{tabular}[c]{@{}c@{}}FedAvg\\ (+IMA)\end{tabular} & \begin{tabular}[c]{@{}c@{}}FedProx\\ (+IMA)\end{tabular} & \begin{tabular}[c]{@{}c@{}}FedASAM\\ (+IMA)\end{tabular} & \begin{tabular}[c]{@{}c@{}}FedFA\\ (+IMA)\end{tabular} & \begin{tabular}[c]{@{}c@{}}FedNova\\ (+IMA)\end{tabular} & \begin{tabular}[c]{@{}c@{}}FedAdam\\ (+IMA)\end{tabular} & \begin{tabular}[c]{@{}c@{}}FedYogi\\ (+IMA)\end{tabular} & \begin{tabular}[c]{@{}c@{}}FedGMA\\ (+IMA)\end{tabular} \\ \midrule
\multirow{2}{*}{\begin{tabular}[c]{@{}c@{}}FMNIST\\ (CNN)\end{tabular}} & $\#C=2$ & 81.17(\textbf{84.68}) & 79.78(\textbf{83.77}) & 84.69(\textbf{85.01}) & \underline{85.60}(\underline{\textbf{88.06}}) & 81.23(\textbf{84.65}) & 83.53(\textbf{86.99}) & 82.42(\textbf{86.86}) & 80.89(\textbf{84.56}) \\
 & $\alpha=0.1$ & 80.13(\textbf{83.06}) & 78.76(\textbf{81.64}) & 80.81(\textbf{82.88}) & \underline{82.97}(\underline{\textbf{86.45}}) & 79.98(\textbf{83.15}) & 78.85(\textbf{83.54}) & 79.66(\textbf{83.95}) & 80.18(\textbf{83.14}) \\ \midrule
\multirow{2}{*}{\begin{tabular}[c]{@{}c@{}}CIFAR-10\\ (CNN)\end{tabular}} & $\#C=2$ & 62.34(\textbf{67.37}) & 61.71(\textbf{67.03}) & 62.60(\textbf{63.64}) & \underline{67.49}(\textbf{69.19}) & 62.34(\textbf{67.46}) & 64.49(\underline{\textbf{69.59}}) & 66.68(\textbf{68.74}) & 62.25(\textbf{67.47}) \\
 & $\alpha=0.1$ & 61.00(\textbf{64.57}) & 61.31(\textbf{64.80}) & 56.92(\textbf{59.10}) & \underline{64.99}(\textbf{\underline{67.03}}) & 55.11(\textbf{60.09}) & 61.61(\textbf{66.25}) & 64.12(\textbf{65.86}) & 61.18(\textbf{64.36}) \\
\multirow{2}{*}{\begin{tabular}[c]{@{}c@{}}CIFAR-10\\ (ResNet)\end{tabular}} & $\#C=2$ & 50.10(\textbf{59.64}) & 53.98(\underline{\textbf{61.65}}) & 49.01(\textbf{56.78}) & 46.56(\textbf{56.15}) & 49.65(\textbf{59.30}) & 54.04(\textbf{59.05}) & \underline{54.45}(\textbf{59.73}) & 49.42(\textbf{58.79}) \\
 & $\alpha=0.1$ & 49.96(\textbf{56.37}) & \underline{52.13}(\textbf{55.07}) & 48.96(\textbf{54.41}) & 42.84(\textbf{48.88}) & 33.72(\textbf{40.52}) & {47.47}(\textbf{47.60}) & 50.92(\textbf{51.26}) & 49.89(\underline{\textbf{55.93}}) \\ \midrule
\begin{tabular}[c]{@{}c@{}}CIFAR-100\\ (VGG)\end{tabular} & \begin{tabular}[c]{@{}c@{}}$\alpha=0.1$\\ (+PA)\end{tabular} & 38.99(\textbf{39.89}) & 38.88(\textbf{39.93}) & 37.51(\textbf{38.25}) & \underline{43.47}(\underline{\textbf{44.68}}) & 39.21(\textbf{39.96}) & 38.96(\textbf{39.83}) & 38.89(\textbf{39.29}) & 39.30(\textbf{40.02}) \\
\begin{tabular}[c]{@{}c@{}}CIFAR-100\\ (ResNet)\end{tabular} & \begin{tabular}[c]{@{}c@{}}$\alpha=0.1$\\ (+PA)\end{tabular} & 31.60(\textbf{32.97}) & 32.06(\textbf{33.27}) & 28.35(\textbf{29.34}) & 31.24(\textbf{34.03}) & 32.01(\textbf{33.50}) & \underline{37.87}(\underline{\textbf{40.93}}) & 37.55(\textbf{40.27}) & 31.65(\textbf{32.90}) \\ \midrule
\multirow{2}{*}{\begin{tabular}[c]{@{}c@{}}Digit Five\\ (CNN)\end{tabular}} &  $\#C=2$(+FS) & 87.90(\textbf{90.15}) & 88.14(\textbf{90.04}) & 88.68(\textbf{89.97}) & \underline{90.26}(\textbf{91.16}) & 87.77(\textbf{89.53}) & 85.63(\underline{\textbf{91.50}}) & 86.31(\textbf{91.25}) & 87.91(\textbf{90.33}) \\
 &  $\alpha=0.1$(+FS) & 90.45(\textbf{91.38}) & 90.52(\textbf{91.48}) & 90.53(\textbf{91.41}) & 90.57(\textbf{91.58}) & 90.10(\textbf{90.76}) & 90.55(\textbf{92.20}) & \underline{91.06}(\underline{\textbf{92.30}}) & 90.50(\textbf{91.49}) \\ \midrule
\multirow{2}{*}{\begin{tabular}[c]{@{}c@{}}PACS\\ (AlexNet)\end{tabular}} &  $\#C=2$(+FS) & 57.47(\textbf{58.01}) & 60.88(\textbf{61.51}) & \underline{{61.15}}(\textbf{61.46}) & 56.57(\textbf{57.36}) & 60.24(\textbf{\underline{63.53}}) & 54.63(\textbf{60.09}) & 55.54(\textbf{57.03}) & {57.33}(\textbf{62.17}) \\
 &  $\alpha=0.1$(+FS) & 40.36(\textbf{47.36}) & \underline{42.15}(\underline{\textbf{49.13}}) & 39.57(\textbf{43.29}) & 41.95(\textbf{47.12}) & 13.96(\textbf{16.10}) & 33.76(\textbf{43.23}) & 39.97(\textbf{40.56}) & 41.73(\textbf{47.46}) \\ \bottomrule
\end{tabular}%
}
\end{table*}
 
\subsubsection{Datasets and Models}
We evaluate the performance of baselines with and without IMA on different models and datasets, considering both label and feature skews. 
Table \ref{table:performance} presents the mean accuracy of the global model for the last ten rounds  (mean top-1 accuracy of all domains in Digit Five and PACS).
For label skew, we train CNN models \cite{mcmahan2017communication} on FMNIST and CIFAR-10, and train ResNet18 \cite{he2016deep} and  VGG11 \cite{simonyan2014very} on CIFAR-10/100.
For label-feature skew, we train CNN on Digit Fives and  AlexNet \cite{krizhevsky2017imagenet} on PACS.
We replace  BN layers with GN layers following \cite{hsieh2020non}.
Detailed settings are presented in Table \ref{table:specific_model} in the Appendix.
We aim to demonstrate the effectiveness of IMA on FL  by considering different model architectures and datasets.

\subsubsection{FL Setup and Baselines}
 In the FL setup, unless otherwise specified, we use a batch size of 50 and 5 local epochs, with  100 clients participating in FL for 400 rounds,  and one-tenth of the clients participate in each round.
 For the client optimizer,  we follow the standard configuration from the FL benchmark \cite{li2021federated} and use the SGD optimizer with a learning rate (lr) of 0.01 and momentum of 0.9   (see Tables \ref{table:hyperparameter}  and \ref{table:testsetup}  for more details in the Appendix).
 
For baselines, in addition to  FedAvg \cite{mcmahan2017communication}, we include other methods that improve FedAvg on the client side, such as parameter-regularization: FedProx \cite{li2020federated}, flatness-improvement: FedASAM \cite{caldarola2022improving}, and feature-classifier-alignment: FedFA \cite{zhou2022fedfa}), and on the server side, such as update-normalization: FedNova \cite{wang2020tackling}, gradient-masking: FedGMA \cite{tenison2022gradient}, and server-momentum: FedADAM/FedYogi \cite{reddi2021adaptive}.
Here, we choose FedAdam and FedYogi as our momentum-based baselines because our method aligns with their approach of using global model updates for momentum, instead of SCAFFOLD \cite{karimireddy2020scaffold}, which relies on receiving updates from a sufficient number of clients in each round and is ineffective when clients have unpredictable availability and may drop out during the training process \cite{li2021federated,sun2023mimic}. 
 Meanwhile, we implement IMA on these baselines with a window size $P=5$ and the starting round $t_s = 0.75R$ with $R=400$, unless otherwise specified. 
It is worth noting that IMA provides a better initialization for client models and is thus compatible with these baselines.

\subsection{Experimental Results}

\subsubsection{Performance with Label Skew} 
Table \ref{table:performance}  illustrates that, for label skew (i.e., $\#C=2$ and $\alpha=0.1$),  IMA enhances the performance of all methods on different datasets and models.
Adding IMA consistently improves performance across all datasets (FMNIST, CIFAR-10, and CIFAR-100). 
For instance, when training a CNN model on FMNIST, FedFA with IMA achieves the highest accuracy of 88.06$\%$  among baselines, compared with 79.7$\%$  for FedProx without IMA.
The most significant improvement is achieved by training ResNet on CIFAR-10, where the performance rises from 49.65$\%$ to 59.30$\%$, i.e., with a gain of 9.65$\%$.
Moreover, for the same setup of label skew, e.g., $\alpha=0.1$, the performance gain for CIFAR-10 is 6.42$\%$  (FedAvg with ResNet), which is twice of the case of CIFAR-100 (3.06$\%$), as shown in Table \ref{table:performance}. Note that CIFAR-100 employs Pachinko Allocation  \cite{li2006pachinko} to make date heterogeneity milder.
Thus, the benefits of IMA depend on the heterogeneity level of label skew, with greater heterogeneity resulting in more significant performance gains.

Meanwhile, the performance of various models on the same dataset varies in FL with the same heterogeneous data setup.
For example, in the case of $\alpha=0.1$ on  CIFAR-10, the CNN model achieves approximately 10\% accuracy improvement over the ResNet model across all methods.
This is partly because the CNN model, with fewer parameters, is faster to train within a given total training round, thereby achieving higher accuracy.
In addition, compared to CNN, the better fitting ability of the ResNet model causes it to overfit heterogeneous local data more significantly after multiple local epochs.
This leads to larger divergence among local models and reduced accuracy of the global model.

\subsubsection{Performance on Various Heterogeneous Degrees}
\begin{figure*}[t]
	\centering  
	\subfigbottomskip=2pt 
	\subfigcapskip=-5pt 
    	\subfigure[Different heterogeneous degrees]{
		\includegraphics[width=0.31\linewidth]{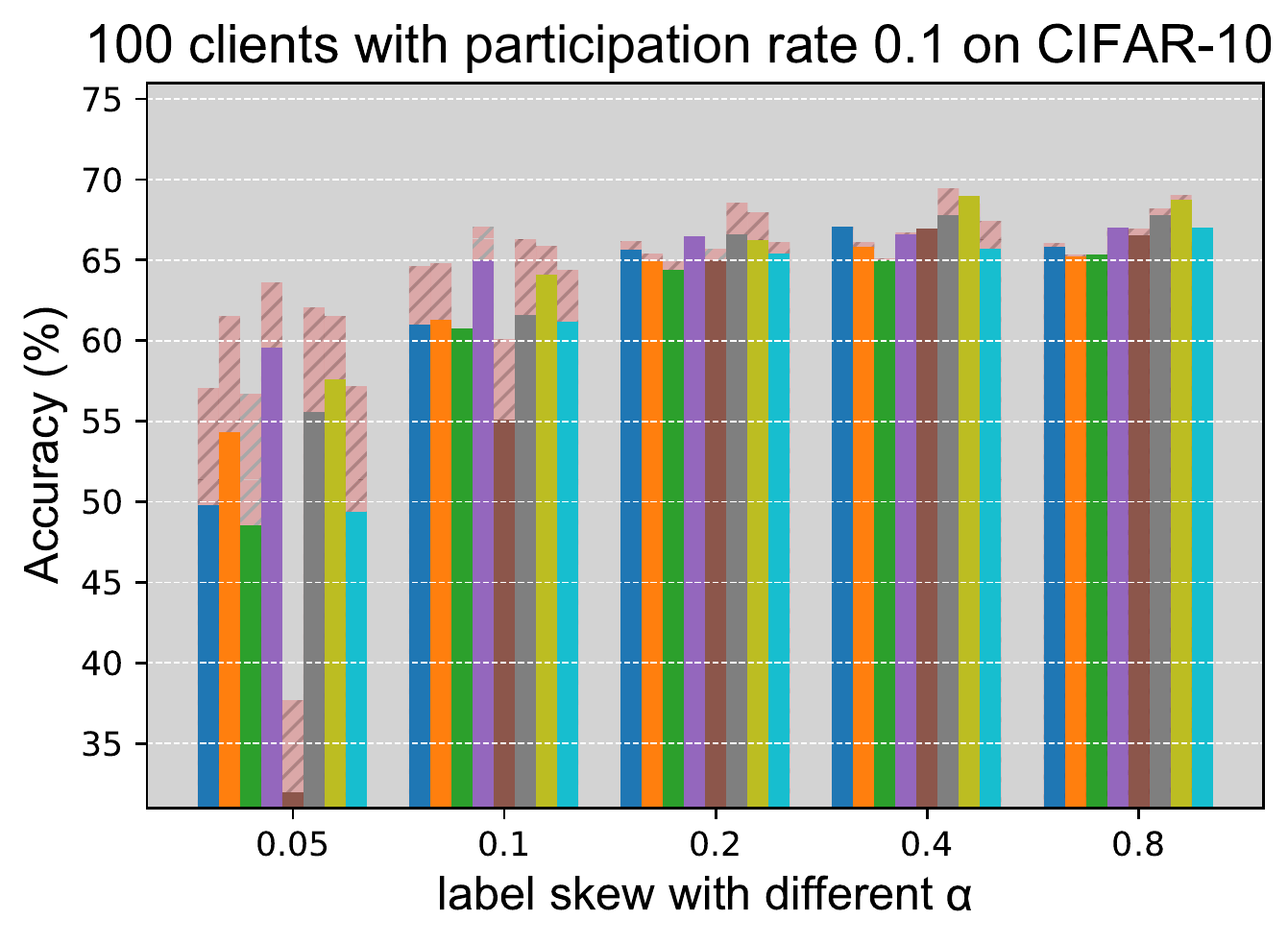}
    \label{fig:different_label_skew}}
	\subfigure[Different participation rates]{
		\includegraphics[width=0.31\linewidth]{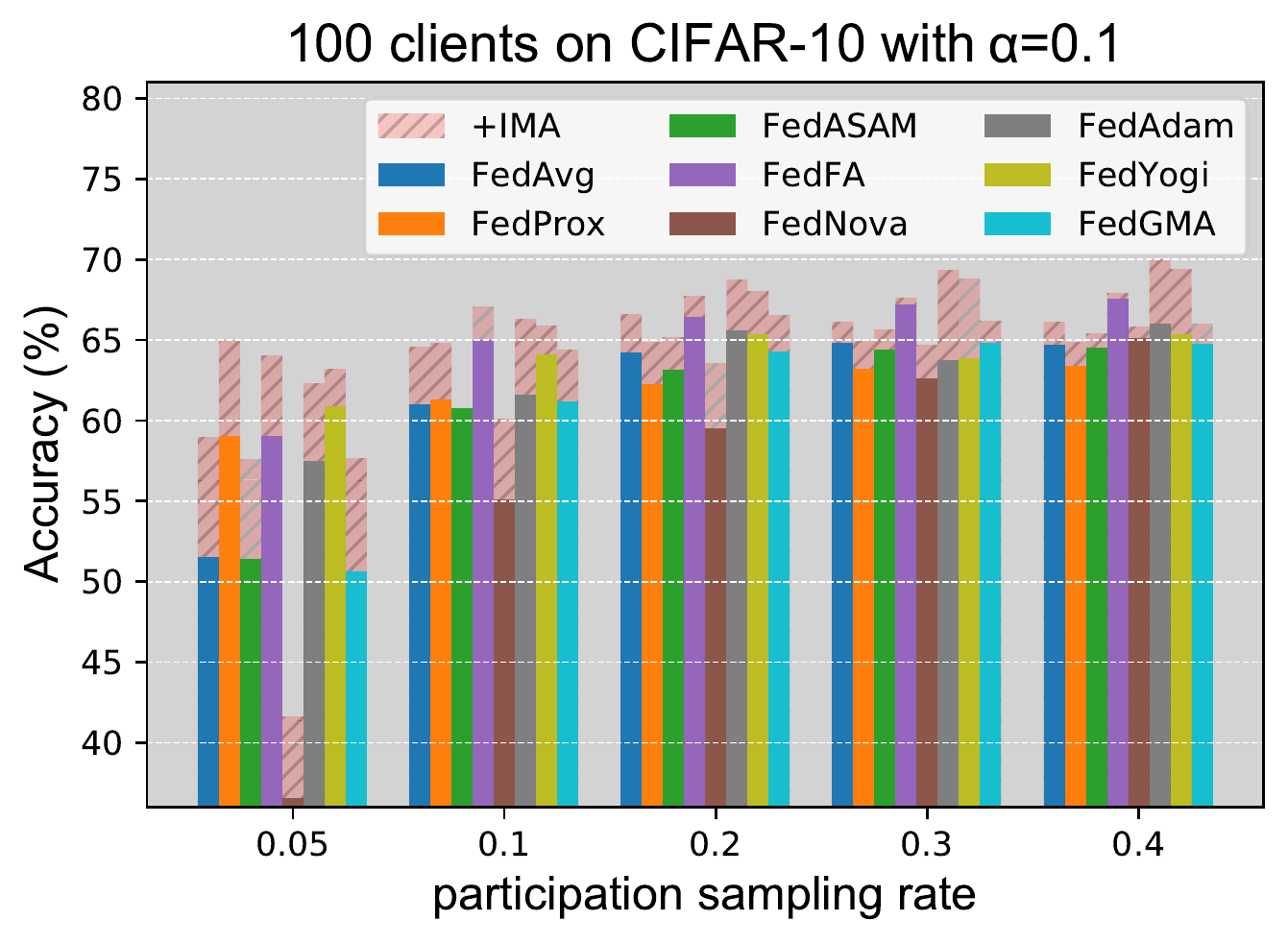}
    \label{fig:different_partip_rate}}
       	\subfigure[Different local epochs]{
		\includegraphics[width=0.31\linewidth]{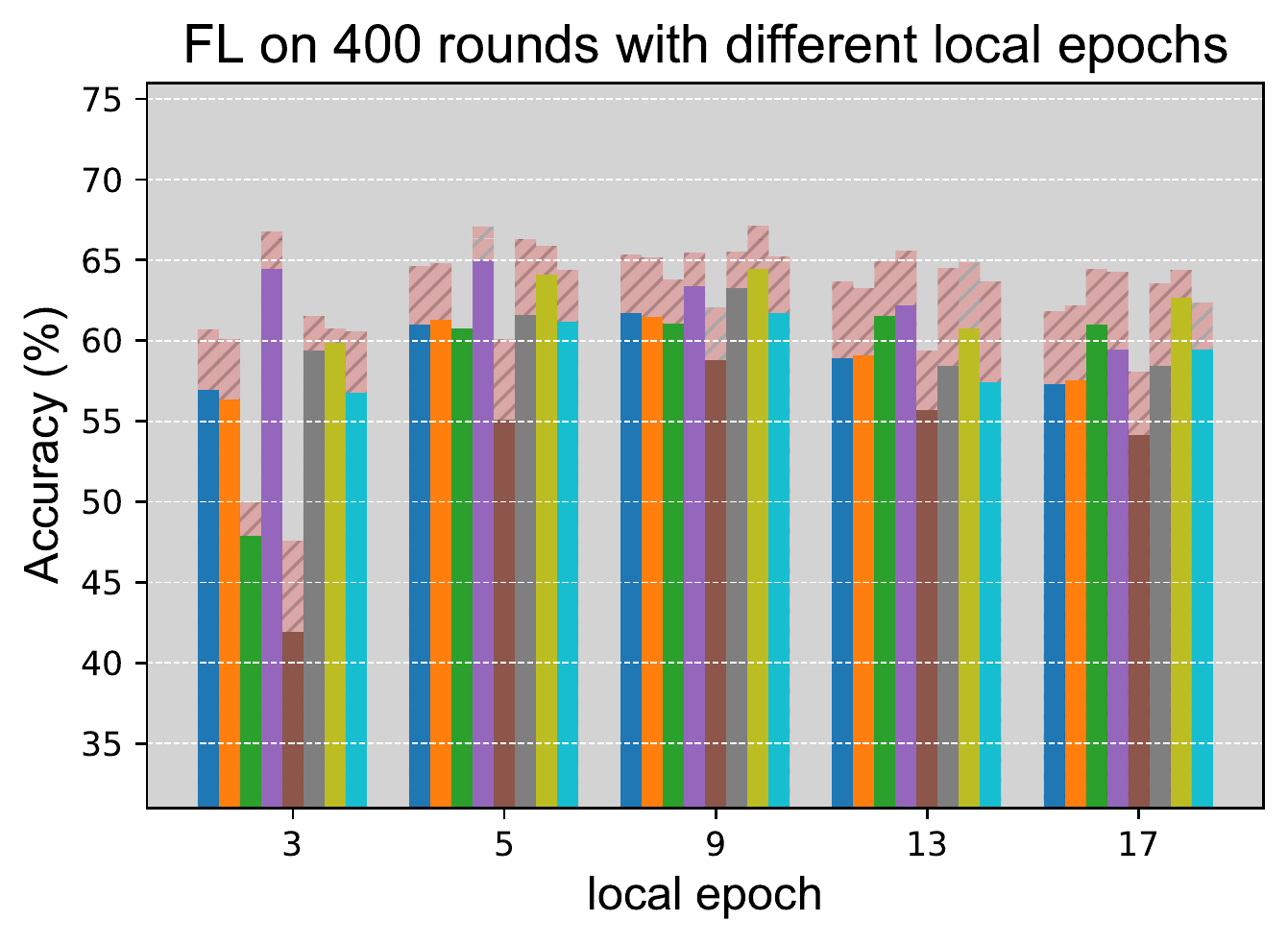}
    \label{fig:different_local_epoch}}
  \caption{Performance of all methods with and without IMA on different federated setups.}
  \label{fig:federated_setups}
\end{figure*}

To further investigate the effect of heterogeneous data on IMA, we conduct tests on Digit Five and PACS datasets under both label and feature skew. 
Our findings on feature skew, as shown in Table \ref{table:performance}, are similar to those observed in the cases of label skew.
For instance, we observe a greater performance gain with IMA on PACS than Digit Five due to the more severe heterogeneity of feature skew in PACS with $\alpha=0.1$. 
To validate these findings, we test different levels of label skew on CIFAR-10 and present the results in Figure \ref{fig:different_label_skew}.
The figure indicates that IMA substantially improves performance on more heterogeneous data, represented by smaller $\alpha$.
Specifically, the IMA gain for $\alpha=0.05$ and $\alpha=0.1$ is approximately 5$\%$ for all baselines, whereas the gain becomes insignificant for $\alpha=0.8$.
Moreover, the performance gain of IMA diminishes as $\alpha$ increases.
This implies that IMA is superior to FMA, except for homogeneous data.
Therefore, Table \ref{table:performance} and Figure \ref{fig:different_label_skew} demonstrate the effectiveness of IMA in mitigating the negative effect of heterogeneous data, especially in scenarios with extreme heterogeneity.

\begin{figure*}[t]
	\centering  
	\subfigbottomskip=2pt 
	\subfigcapskip=-2pt 
    	\subfigure[Different start round $t_s$]{
		\includegraphics[width=0.3\textwidth]{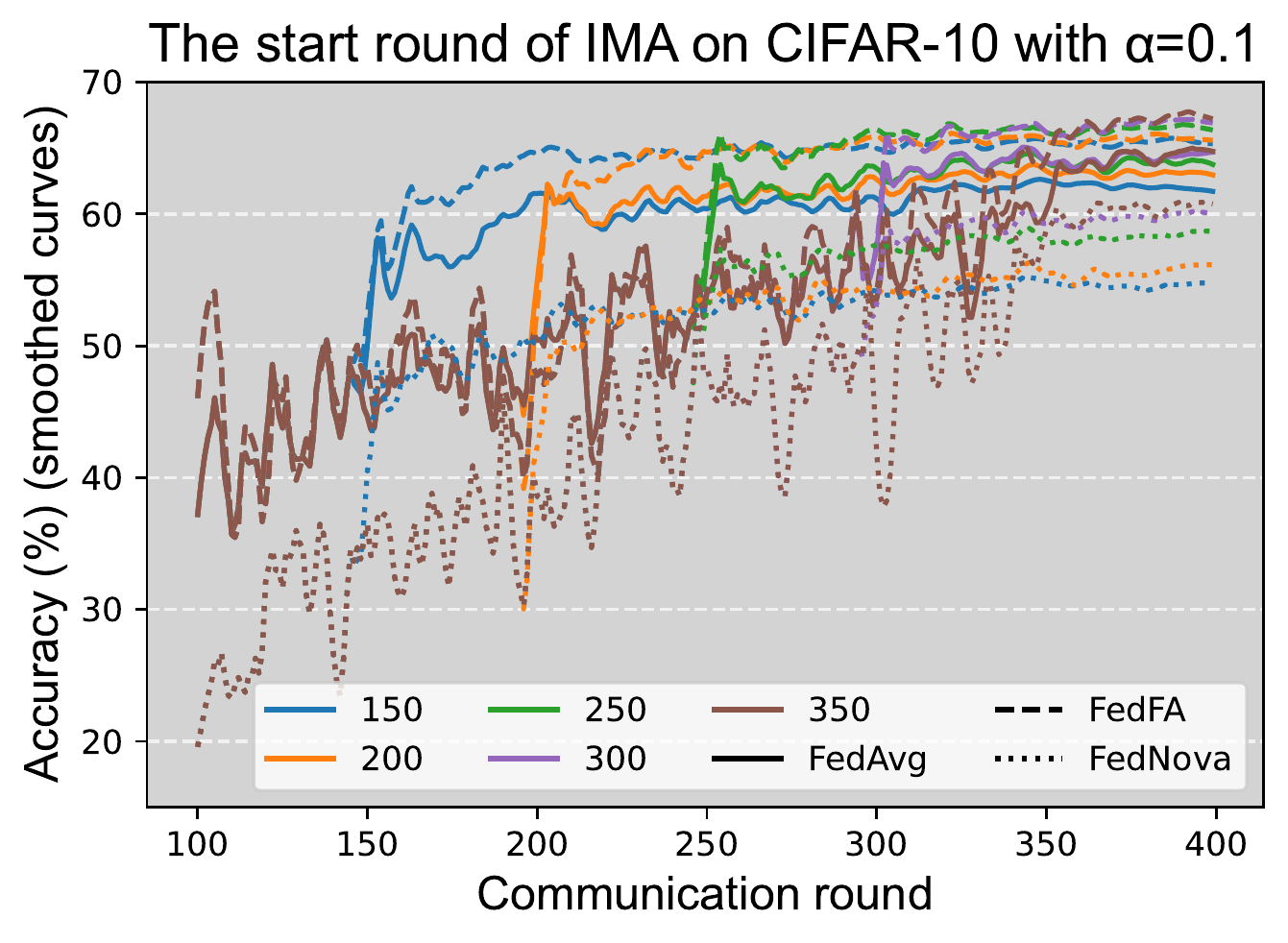}
    \label{fig:start_round}}
	\subfigure[Different window size $P$]{
		\includegraphics[width=0.3\textwidth]{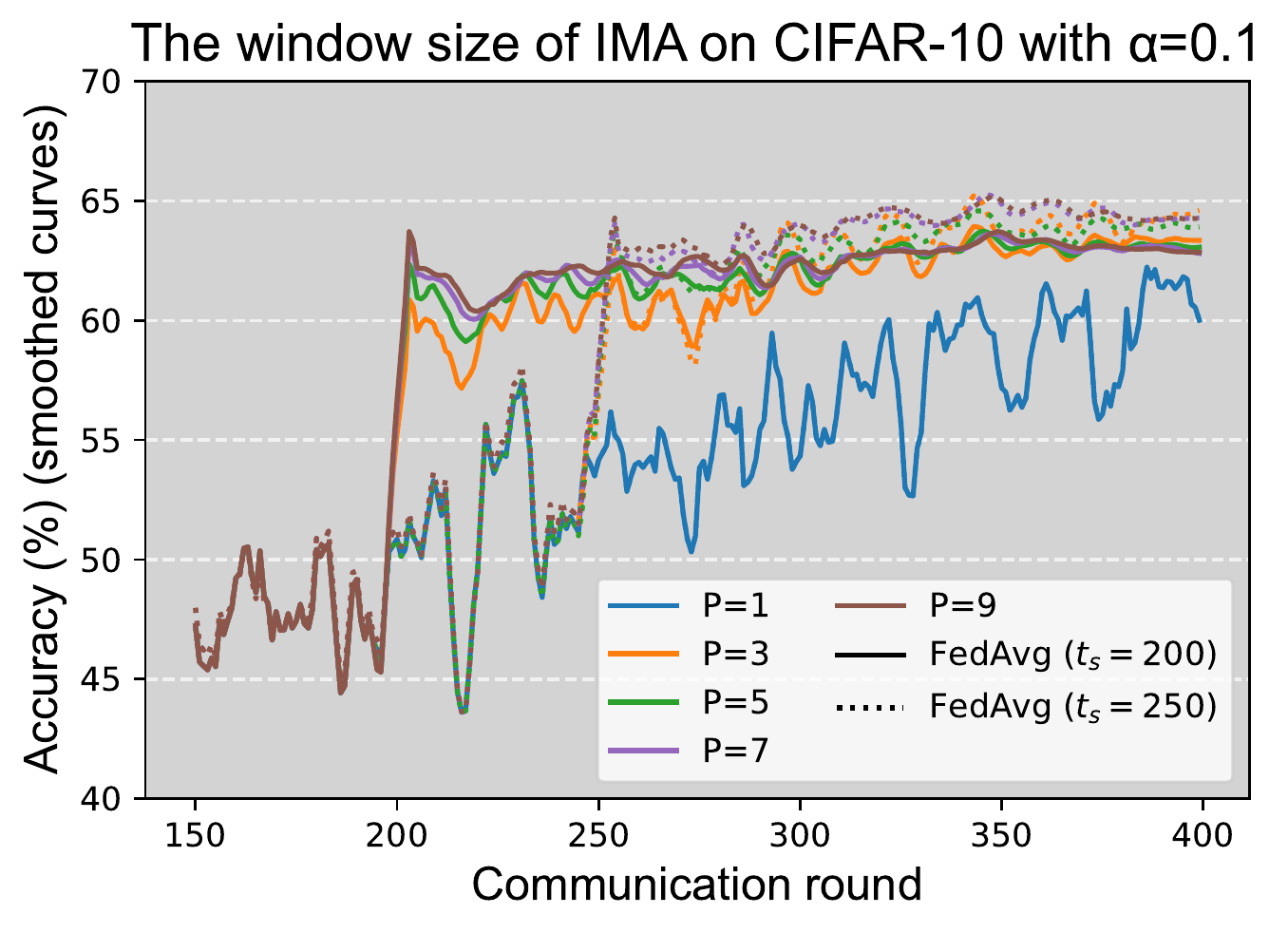}
    \label{fig:window_size}}
    	\subfigure[Test error/loss landscape]{
		\includegraphics[width=0.325\textwidth]{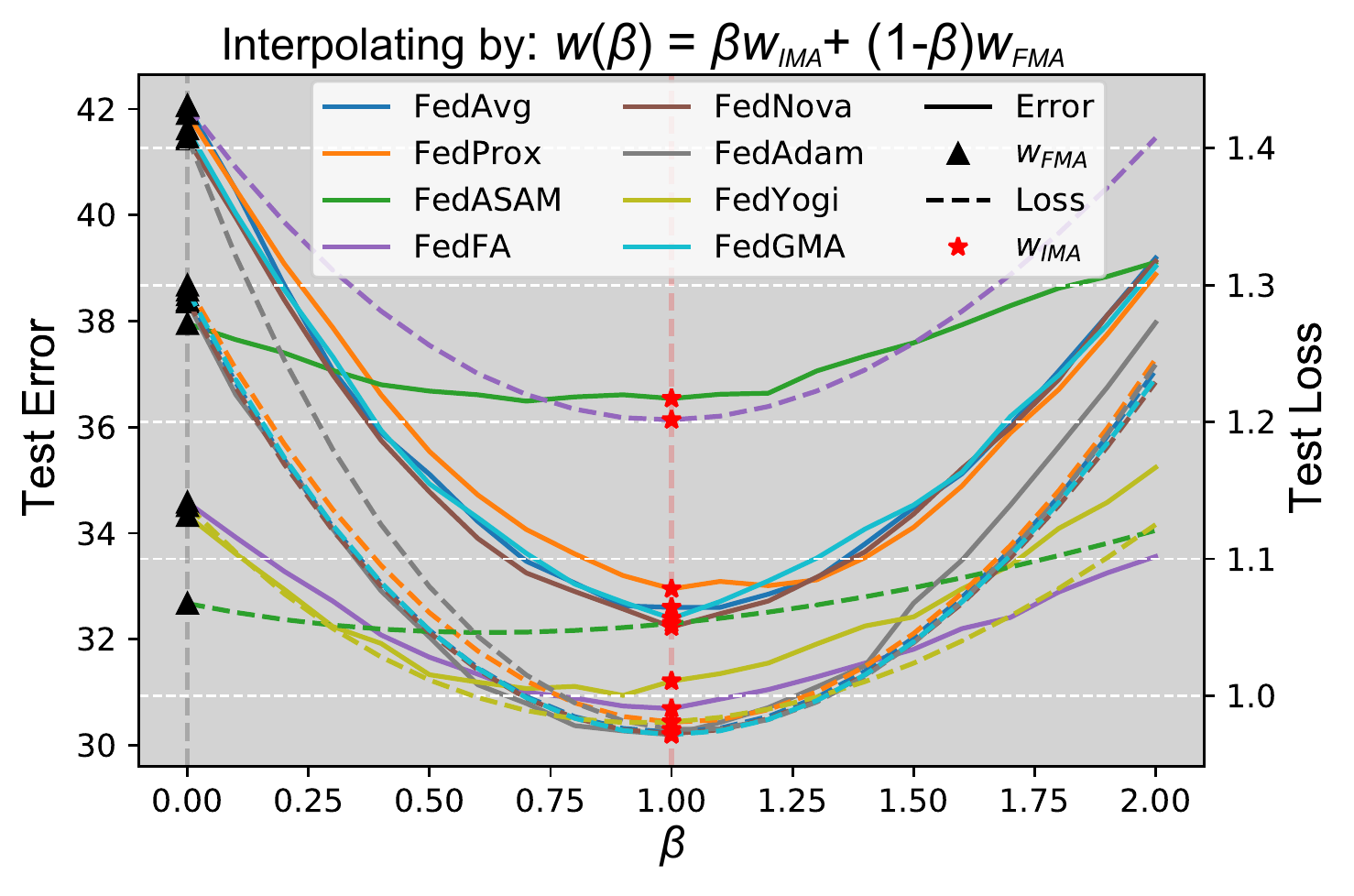}
    \label{fig:loss_landscape_two_model}}
  \caption{Ablation studies on IMA and landscape visualization between FMA and IMA models. 
 Note that all curves in Figures \ref{fig:start_round} and \ref{fig:window_size} are smoothed by a Savitzky-Golay filter  \cite{Schafer112011} with a window length of 10 and a polynomial order of 2 to mitigate the noise in the visualization of the results while preserving the essential trends.  }
  \label{fig:ablation_study}
\end{figure*}

\subsubsection{Reduction in Communication Overhead}
Table \ref{table:Communication_effeciency} presents the communication efficiency of IMA with different starting rounds to achieve a target accuracy on CIFAR-10 with $\alpha=0.1$, where FedASAM and FedNova are not reported because their performance is worse than the targeted accuracy.
The results illustrate that initiating IMA at earlier rounds significantly reduces the communication overhead, compared with three-quarters of the total rounds in Table \ref{table:performance}.
 For instance, starting IMA at the 150th round saves communication by nearly half for FedAdam and   FedProx.
 
 \begin{table}[t]
\centering
\caption{Required rounds by IMA with different start rounds $t_s$ when the accuracy reaches $61.61\%$ from Table \ref{table:performance}.}
\label{table:Communication_effeciency}
\resizebox{0.5\textwidth}{!}{%
\begin{tabular}{cccccl}
\hline
(+IMA) $t_s$ & 150 & 200 & 250 & 300 & FMA \\ \hline
FedAvg & 318($\times$1.24) & 257($\times$1.53) & 260($\times$1.51) & 309($\times$1.28) & 394($\times$1) \\
FedProx & 201($\times$1.96) & \textbf{210}($\times$\underline{1.88}) & 260($\times$\underline{1.52}) & 309($\times$1.28) & 394($\times$1) \\
FedFA & \textbf{183}($\times$1.87) & 212($\times$1.61) & 258($\times$1.32) & 308($\times$1.11) & 341($\times$1) \\
FedAdam & 193($\times$\underline{2.03}) & \textbf{210}($\times$1.87) & 259($\times$1.51) & \textbf{306}($\times$\underline{1.28}) & 392($\times$1) \\
FedYogi & 195($\times$1.85) & \textbf{210}($\times$1.71) & \textbf{257}($\times$1.40) & \textbf{306}($\times$1.18) & 360($\times$1) \\
FedGMA & 316($\times$1.25) & 242($\times$1.63) & 260($\times$\underline{1.52}) & 309($\times$1.28) & 394($\times$1) \\ \hline
\end{tabular}%
}
\end{table}

\subsubsection{Performance on Different Client Participation Rates}
We evaluate the performance of IMA under varying participation rates from $0.05$ to $0.4$ in Figure \ref{fig:different_partip_rate}, in addition to the results obtained with a $0.1$ participation rate in Table \ref{table:performance}.
The figure indicates that the gain achieved by IMA generally increases as the client participation rate decreases. For example, the gain with a $0.05$ participation rate is approximately twice that observed with a $0.2$ participation rate.
Furthermore, Figure \ref{fig:different_partip_rate} verifies the global model deviation induced by low participation rates, as highlighted in Section \ref{section:proposed_method}. 
It illustrates that lower participation rates lead to larger deviations between the cohort and the global datasets, amplifying the negative effect of heterogeneous data.

\subsubsection{Performance on Different Local Epochs}
To assess the robustness of IMA, we evaluate its performance on different local epoch settings ranging from 3 to 17, as shown in Figure \ref{fig:different_local_epoch}.
The results show that IMA consistently improves all baseline methods across different epochs.
 We also observe that the performance gain remains stable even when the number of local epochs increases.
  This is because client models are closely located around the global model within the same basin due to FMA, as observed in Section \ref{section:loss_landscape}.
Consequently, the advantages of IMA persist even when client models are close to their local optima, as IMA may bring global models closer to the global optimum.

\subsection{Ablation Study on IMA}
\subsubsection{Ablation on Starting Rounds and Window Size of IMA}
The results presented in Table \ref{table:Communication_effeciency} indicate that initiating IMA at a later round leads to increased communication overhead when considering a target accuracy. For example, setting $t_s=300$ on the FedFA baseline results in an additional 96 rounds compared to the case of $t_s=200$.
In contrast, Figure \ref{fig:start_round} demonstrates that starting IMA at a later round leads to better accuracy performance. For instance, the case of $t_s=300$ on FedFA shows an approximately 3\% increase in accuracy compared to the case of $t_s=200$.
These findings highlight the existence of a trade-off between communication efficiency and performance in IMA.
Moreover, increasing the window size improves the training stability, but it impairs the final accuracy if IMA starts early.
This can be observed in the case of FedAvg with $t_s=200$ and $P=9$, where a lower accuracy is achieved compared to other cases, as shown in Figure \ref{fig:window_size}.
Note that due to the oscillation of the original results, all curves in Figures \ref{fig:start_round} and  \ref{fig:window_size}  have been smoothed for better visualization clarity.

 \begin{table}[t]
\centering
\caption{Accuracy v.s. decay schemes.}
\label{table:decaying_schemes}
\resizebox{0.4\textwidth}{!}{%
\begin{tabular}{@{}cccccc@{}}
\toprule
\begin{tabular}[c]{@{}c@{}}IMA w/\\ decay\end{tabular} & \begin{tabular}[c]{@{}c@{}}Exp\\ Decay\end{tabular} & \begin{tabular}[c]{@{}c@{}}Const\\ LR\end{tabular} & \begin{tabular}[c]{@{}c@{}}Cyclic\\ Decay\end{tabular} & \begin{tabular}[c]{@{}c@{}}Epoch\\ Decay\end{tabular} & \begin{tabular}[c]{@{}c@{}}NA\\ Decay\end{tabular} \\ \midrule
FedAvg & \textbf{64.57} & 64.50 & 64.27 & 62.96 & 63.96 \\
FedProx & \textbf{64.80} & 64.73 & 64.59 & 63.14 & 64.12 \\
FedASAM & 59.10 & 58.14 & \textbf{59.33} & 57.48 & 58.95 \\
FedFA & \textbf{67.03} & 66.62 & 66.94 & 66.63 & 66.40 \\
FedNova & \textbf{60.09} & 59.86 & 59.38 & 59.04 & 58.90 \\
FedAdam & \textbf{66.25} & 65.89 & 66.06 & 64.00 & 65.62 \\
FedYogi & \textbf{65.86} & 65.53 & 65.51 & 63.13 & 65.12 \\
FedGMA & 64.36 & \textbf{64.41} & 64.12 & 62.97 & 63.75 \\ \bottomrule
\end{tabular}%
}
\end{table}

\subsubsection{Ablation on Mild Client Exploration in IMA}
As mentioned in Section \ref{section:proposed_method},    we adopt a more aggressive exponential lr decay per round in IMA than in FMA to restrict client exploration. 
To evaluate this design choice, we conduct experiments on CIFAR-10 with $\alpha=0.1$ to ablate IMA with different decay schemes, including a small constant lr (i.e., lr is $5\times10^{-5}$ in IMA), cyclic lr decay \cite{izmailov2018averaging} (i.e., decaying lr from $1\times10^{-2}$ to $5\times10^{-5}$ every 20 rounds), epoch decay \cite{pu2021server} (i.e., decaying one local epoch per 20 rounds), and non-additional decay (NA).
As shown in Table \ref{table:decaying_schemes}, more aggressive decay schemes that sufficiently constrain client updates (e.g.,  exponential lr decay or small constant lr) outperform milder schemes.
For instance, exponential decay achieves $64.57\%$ on FedAvg, compared with $62.96 \%$ of epoch decay.

\subsubsection{Test Loss Landscape between FMA and IMA Models}
 Figure \ref{fig:loss_landscape_two_model} depicts the interpolation model between FMA and IMA models (both from the final round) to visualize the landscape of test error and test loss.
The figure shows that the IMA models reach almost the center (i.e., the lowest point) of the test error and loss basins for all baselines, effectively alleviating the deviation mentioned in Section \ref{section:proposed_method}. 
In contrast, the FMA models only reach the basin's wall, which verifies the deviation observed in Section \ref{section:loss_landscape}.
Moreover, Figure \ref{fig:loss_landscape_two_model} shows that these methods reach various basins with different curvature. 
However, it does not necessarily hold that a flatter basin corresponds to lower error. 
For example, while FedASAM reaches the basin with the flattest curvature, it achieves the highest test error.

\section{Discussions and Future works}\label{section:conclusion}
This work advanced the understanding of how FMA operates in the presence of heterogeneous data and proposed employing IMA to enhance its performance. 
Firstly, we investigated the dynamics of the loss landscape of FMA during training and observed that client models closely surround the global model within the same basin.
By employing test loss decomposition, we illustrated the relationship between the global model and client models, demonstrating that the client models' heterogeneous bias and locality dominate the global model's error after the early training stage.
These findings motivated us to adopt IMA on global models in the late training stage rather than disregarding them in FMA.
Our experiments showed that IMA  significantly improves existing FL methods' accuracy and communication efficiency under both label and feature skews.

Although we demonstrate the error relationship between the global model and client models based on expected loss decomposition in Section \ref{section:loss decom}, it remains necessary to explicitly quantify this relationship in general cases. 
Future works should analyze how each factor dominates the error throughout the training process.  
In addition, an IMA variant with an adaptive starting round demonstrates promising results in Table \ref{table:Communication_effeciency} and deserves investigation to reduce communication overhead without compromising generalization.
Moreover, employing more flexible regularization between the global model and client models (e.g., elastic weight consolidation \cite{kirkpatrick2017overcoming}) can further reduce the bias and locality in Theorem \ref{theorem1}.
We hope our study will serve as a valuable reference for further analysis and improvement of FL methods.

\bibliography{main}
\bibliographystyle{IEEEtran}
\begin{IEEEbiography}[{\includegraphics[width=1in,height=1.25in,clip,keepaspectratio]{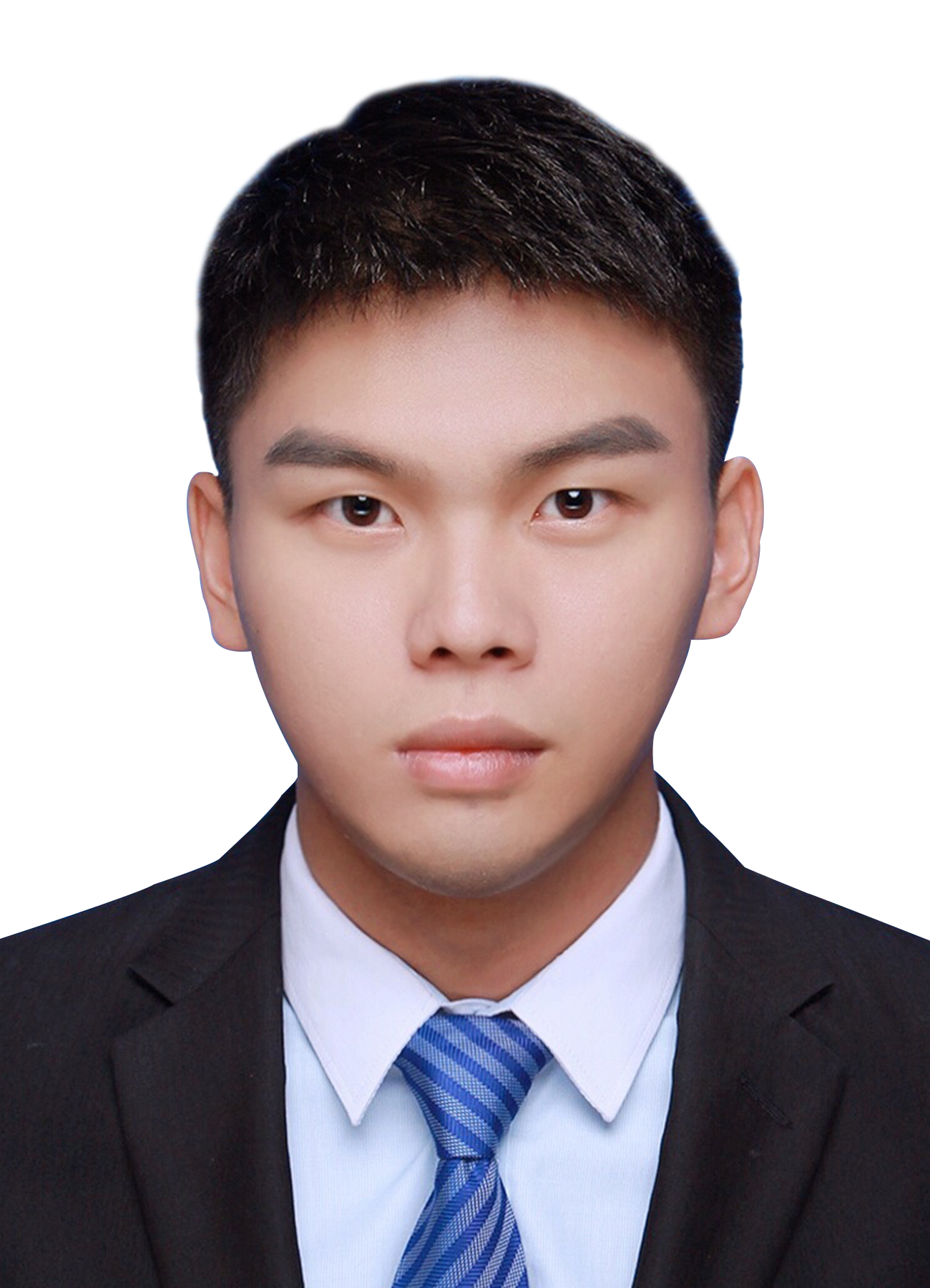}}]{Tailin Zhou}
(Graduate student member, IEEE) received his B.Eng. degree in Electrical Engineering and Automation Engineering from Sichuan University in 2018, and his Master's degree in Electrical Engineering from South China University of Technology in 2021.
He is pursuing a Ph.D. degree at the Hong Kong University of Science and Technology under the supervision of  Professor Jun Zhang and Professor Danny H.K. Tsang.
His research interests include federated learning and cooperative AI agents.
\end{IEEEbiography}
\vskip -2\baselineskip plus -1fil

\begin{IEEEbiography}[{\includegraphics[width=1in,height=1.25in,clip,keepaspectratio]{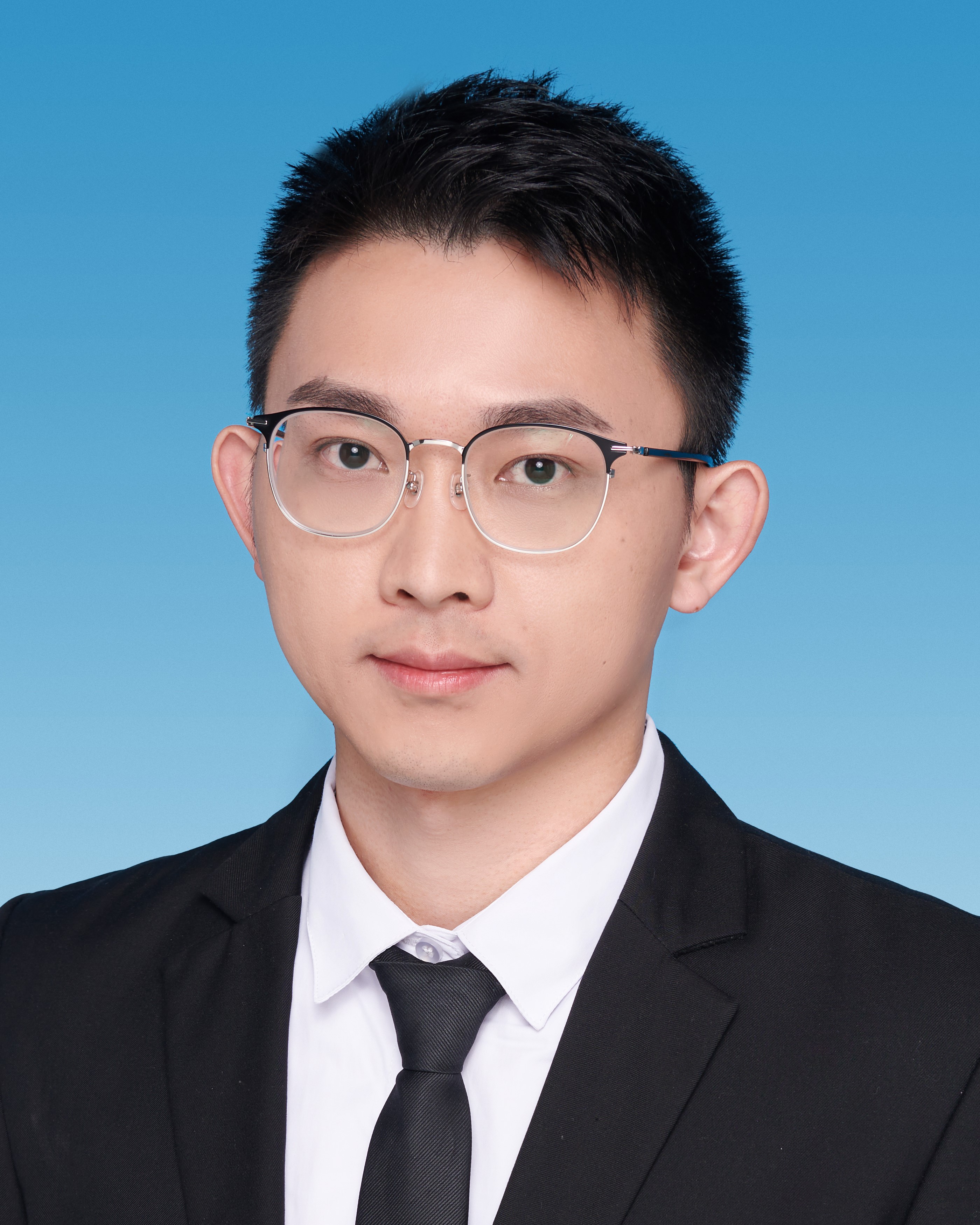}}]{Zehong Lin}
(Member, IEEE) received the B.Eng. degree in information engineering from South China University of Technology in 2017, and the Ph.D. degree in information engineering from The Chinese University of Hong Kong in 2022. Since 2022, he has been with the Department of Electronic and Computer Engineering at the Hong Kong University of Science and Technology, where he is currently a Research Assistant Professor. His research interests include federated learning and edge AI.
\end{IEEEbiography}
\vskip -2\baselineskip plus -1fil

\begin{IEEEbiography}
[{\includegraphics[width=1in,height=1.25in,clip,keepaspectratio]{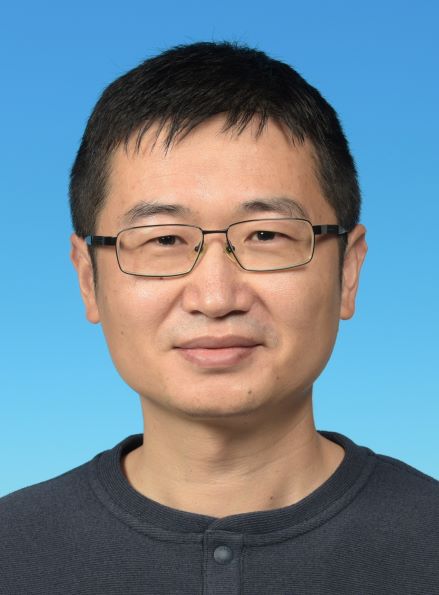}}]{Jun Zhang}
(Fellow, IEEE) received the B.Eng. degree in Electronic Engineering from the University of Science and Technology of China in 2004, the M.Phil. degree in Information Engineering from the Chinese University of Hong Kong in 2006, and the Ph.D. degree in Electrical and Computer Engineering from the University of Texas at Austin in 2009. He is an Associate Professor in the Department of Electronic and Computer Engineering at the Hong Kong University of Science and Technology. His research interests include wireless communications and networking, mobile edge computing and edge AI, and cooperative AI.

Dr. Zhang co-authored the book Fundamentals of LTE (Prentice-Hall, 2010). 
He is an Editor of IEEE Transactions on Communications, IEEE Transactions on Machine Learning in Communications and Networking, and was an editor of IEEE Transactions on Wireless Communications (2015-2020).
He was a MAC track co-chair for IEEE Wireless Communications and Networking Conference (WCNC) 2011 and a co-chair for the Wireless Communications Symposium of IEEE International Conference on Communications (ICC) 2021. He is an IEEE Fellow and an IEEE ComSoc Distinguished Lecturer.

\end{IEEEbiography}  
\vskip -2\baselineskip plus -1fil

\begin{IEEEbiography}
[{\includegraphics[width=1in,height=1.25in,clip,keepaspectratio]{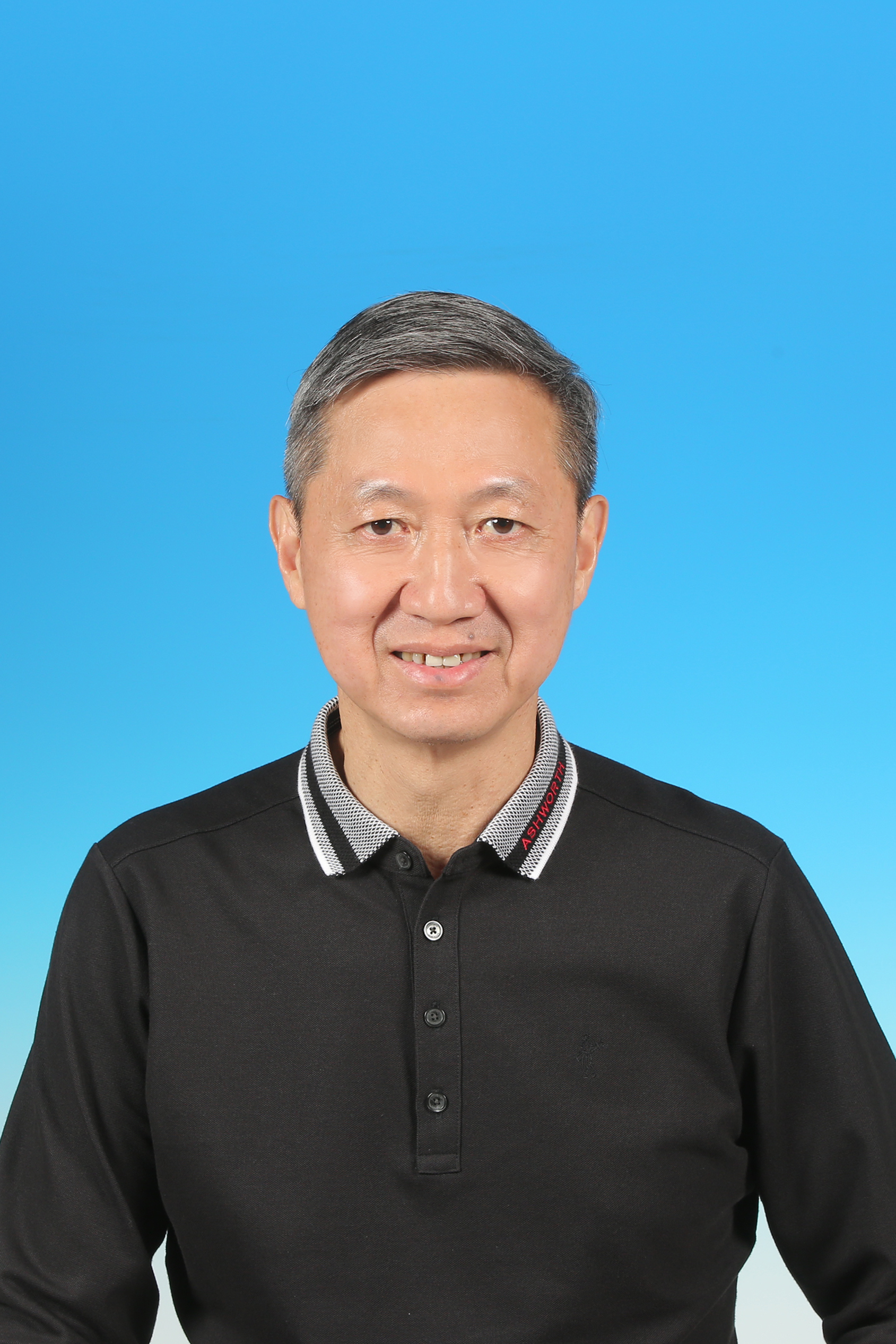}}]{Danny H.K. Tsang}
(Life Fellow, IEEE)  received the Ph.D. degree in electrical engineering from the Moore School of Electrical Engineering, University of Pennsylvania, Philadelphia, PA, USA, in 1989. After graduation, he joined the Department of Computer Science, Dalhousie University, Halifax, NS, Canada. He later joined the Department of Electronic and Computer Engineering, The Hong Kong University of Science and Technology (HKUST), Hong Kong, in 1992, where he is currently a Professor. He has also been serving as the Thrust Head of the Internet of Things Thrust, HKUST (Guangzhou), Guangzhou, China, since 2020. 
  His current research interests include next-generation networking, mobile edge computing, online algorithm design, and smart grids.
  
Dr. Tsang is a member of the Special Editorial Cases Team of IEEE Communications Magazine. 
He was a Guest Editor of the IEEE JOURNAL ON SELECTED AREAS IN COMMUNICATIONS’ special issue on Advances in P2P Streaming Systems, an Associate Editor of Journal of Optical Networking published by the Optical Society of America, and a Guest Editor of IEEE SYSTEMS JOURNAL.  
He invented the 64B/65B encoding (U.S. Patent No.: U.S. 6 952 405 B2) and contributed it to the proposal for Transparent GFP in the T1X1.5 standard that was advanced to become the ITU G.GFP standard. 
The coding scheme has now been adopted by International Telecommunication Union (ITU)’s Generic Framing Procedure Recommendation GFP-T (ITUT G.7041/Y.1303) and Interfaces of the Optical Transport Network (ITU-T G.709). 
He has been elevated to an IEEE Fellow in 2012 and an HKIE Fellow in 2013. 
\end{IEEEbiography}   

\clearpage
\appendix
\section*{Appendix A: Proof}\label{appendix:proof}
\subsection*{Proof of Lemma \ref{lemma1}}\label{Proof_Lemma1}
Suppose clients do not have an extremely imbalanced dataset (i.e., $n_i/n_j \neq \infty$ when $i\neq j$).
 For client $k$ and the FMA's model, we have:
  \begin{equation}
\begin{aligned}
f_{\mathbf{w}_{k}}(x) = f_{\mathbf{w}_{\rm FMA}}(x) +  \langle \Delta f_{\mathbf{w}_{\rm FMA}}(x), \Delta_k   \rangle + O(\|\Delta_k\|^2 )
\end{aligned},
\end{equation}
where $ \langle \cdot , \cdot   \rangle $ is the dot product, and $\Delta_k = \mathbf{w}_{k} -\mathbf{w}_{\rm FMA} $.
Thus, we establish the relationship between the FMA-model function and the WENS function as:
  \begin{equation}
\begin{aligned}
&f_{\rm WENS}(x)   =  \sum_{k=1}^K \frac{n_k}{n} f_{\mathbf{w}_{k}}(x) \\
  =&  f_{\mathbf{w}_{\rm FMA}}(x) + \sum_{k=1}^K \frac{n_k}{n}   \langle \Delta f_{\mathbf{w}_{\rm FMA}}(x), \Delta_k   \rangle + \sum_{k=1}^K \frac{n_k}{n}  O(\|\Delta_k\|^2 )\\
=&  f_{\mathbf{w}_{\rm FMA}}(x) +  \langle \Delta f_{\mathbf{w}_{\rm FMA}}(x),  \sum_{k=1}^K \frac{n_k}{n}  \Delta_k   \rangle +  O(\|\Delta\|^2 )\\
 = & f_{\mathbf{w}_{\rm FMA}}(x) +  O(\|\Delta\|^2 ),
\end{aligned}
\end{equation}
where $\|\Delta\|=\max_k^K\|\Delta_k\|$,  $n= \sum_k n_k$ is the total sample number.
\subsection*{Proof of Theorem \ref{lemma1}}\label{Proof_theorem1}
Substituting $f_{\rm WENS}(x)= \sum_{k=1}^K \frac{n_k}{n} f_{\mathbf{w}_k}$ into (\ref{decomposistion:step1}), we have:
\begin{equation}
    \begin{aligned}
&\mathbb{E}_{ \mathbf{\{w}_k\}_{k=1}^K \in \prod_{k}^K \mathcal{W}_{\mathcal{D}_{k}}} \mathcal{L} ({ \mathbf{\{w}_k\}_{k=1}^K})
 \\ = & \mathbb{E}_{(x,y) \in \mathcal{D}} [( {\rm{Bias}}  \{f_{\rm WENS}|(x,y)\})^2 
   + {\rm{Var}}  \{f_{\rm WENS}|x\}  ].
    \end{aligned}
    \label{decomposistion:step2}
\end{equation}
 \textbf{For the bias term}, we have:
\begin{equation}
    \begin{aligned}
& {\rm{Bias}}  \{f_{\rm WENS}|(x,y)\}  =
  y- \mathbb{E}_{ \mathbf{\{w}_k\}_{k=1}^K} f_{\rm WENS} (x) 
  \\ =&  
 y- \mathbb{E}_{ \mathbf{\{w}_k\}_{k=1}^K} [ \sum_{k=1}^K \frac{n_k}{n}   f_{\mathbf{w}_k}(x)] \\
=  &   y- \sum_{k=1}^K \frac{n_k}{n}   \mathbb{E}_{ \mathbf{w_k}} [  f_{\mathbf{w}_k}(x)] \\= &   \sum_{k=1}^K \frac{n_k}{n} (  y- \mathbb{E}_{ \mathbf{w_k}} [  f_{\mathbf{w}_k}(x)] ).
    \end{aligned}
    \label{decomposistion:step3}
\end{equation}
Taking the expectation of the bias term for the global dataset, we have:
 \begin{equation}
    \begin{aligned}
 & \mathbb{E}_{(x,y) \in \mathcal{D}} ({\rm{Bias}}  \{f_{\rm WENS}|(x,y)\})^2 \\
=   & \frac{1}{n} \sum_{(x,y)\in\mathcal{D}}
   [ 
  \sum_{k=1}^K \frac{n_k}{n}  \underbrace{ \mathbb{I}[{(x,y)\in\mathcal{D}_k}] (y- \mathbb{E}_{ \mathbf{w_k}} [  f_{\mathbf{w}_k}(x)] )}_{{\rm{TrainBias}} \{f_{\mathbf{w}_k}|(x,y)\}} 
 \\  &   +   \sum_{k=1}^K \frac{n_k}{n}  \underbrace{ \mathbb{I} [{(x,y)\in \mathcal{D} \setminus \mathcal{D}_k} ]   (y- \mathbb{E}_{ \mathbf{w_k}} [  f_{\mathbf{w}_k}(x)]  )}_{{\rm{HeterBias}} \{f_{\mathbf{w}_k}|(x,y)\}} 
    ]^2 .
    \end{aligned}
    \label{decomposistion:step4}
\end{equation}
\textbf{For the variance term}, we have:
\begin{equation}
    \begin{aligned}
    &  {\rm{Var}}  \{f_{\rm WENS}|(x,y)\}  
 \\ =  &     \mathbb{E}_{ \mathbf{\{w}_k\}_{k=1}^K} 
     [  (
    \sum_{k=1}^K \frac{n_k}{n}  f_{\mathbf{w}_k}(x)
    - \mathbb{E}_{ \mathbf{\{w}_k\}_{k=1}^K} 
     [   \sum_{k=1}^K \frac{n_k}{n}  f_{\mathbf{w}_k}(x)
     ] )^2
     ] 
    \\ =  &   \sum_{k=1}^K \frac{n_k^2}{n^2}  \underbrace{  \mathbb{E}_{ \mathbf{{w}_k} }
     [ (
    f_{\mathbf{w}_k}(x) - \mathbb{E}_{ \mathbf{{w}_k} }[f_{\mathbf{w}_k}(x)]
     )^2 ]}_{{\rm Var}\{f_{\mathbf{w}_k}|x\}}   
    \\ &   +  \sum_k \sum_{k\neq {k^\prime}} \frac{n_k n_{{k^\prime}}}{n^2} {{\rm Cov}\{f_{\mathbf{w}_k,\mathbf{w}_{{k^\prime}}}|x\}},
    \end{aligned}
    \label{decomposistion:step5}
\end{equation}
where ${\rm Cov}\{f_{\mathbf{w}_k,\mathbf{w}_{{k^\prime}}}|x\} =  \mathbb{E}_{ \mathbf{{w}_k},\mathbf{{w}_{{k^\prime}}}}
    [ 
     (f_{\mathbf{w}_k}(x) - \mathbb{E}_{ \mathbf{{w}_k} }[f_{\mathbf{w}_k}(x)] ) $ $
     (f_{\mathbf{w}_{{k^\prime}}}(x) - \mathbb{E}_{ \mathbf{{w}_{{k^\prime}}} }[f_{\mathbf{w}_{{k^\prime}}}(x)] )
    ]$.
Taking  the expectation of the variance term for the global dataset, we have:
\begin{equation}
    \begin{aligned}
& \mathbb{E}_{(x,y) \in \mathcal{D}} ( {\rm{Var}}  \{f_{\rm ENS}|(x,y)\})
\\= & \mathbb{E}_{(x,y) \in \mathcal{D}}  ( { \sum_{k=1}^K \frac{n_k^2}{n^2} {\rm Var}\{f_{\mathbf{w}_k}|x\}}  
\\ & + {\sum_k \sum_{{k^\prime}} \frac{n_k n_{{k^\prime}}}{n^2}  {\rm Cov}\{f_{\mathbf{w}_k,\mathbf{w}_{k^\prime}}|x\}  } )
\\= & \frac{1}{n} \sum_{(x,y)\in\mathcal{D}}  { \sum_{k=1}^K \frac{n_k^2}{n^2} {\rm Var}\{f_{\mathbf{w}_k}|x\}}  
\\ & + \frac{1}{n} \sum_{(x,y)\in\mathcal{D}}  {\sum_k \sum_{{k^\prime}} \frac{n_k n_{{k^\prime}}}{n^2}  {\rm Cov}\{f_{\mathbf{w}_k,\mathbf{w}_{k^\prime}}|x\}}.
    \end{aligned}
    \label{decomposistion:step6}
\end{equation}

Using the   Taylor expansion at the zeroth order of the loss, we extend Lemma \ref{lemma1} and obtain:
\begin{equation}
    \begin{aligned}
 & \mathcal{L} (\mathbf{{w}_{\rm FMA}} ) =     \mathbb{E}_{(x, y) \in \mathcal{D}}[  l(f_{\mathbf{{w}_{\rm FMA}}}(x); y)] 
 \\= &    \mathbb{E}_{(x, y) \in \mathcal{D}}[  l(f_{\rm WENS}(x); y)] 
 \\&  + O(\|f_{\mathbf{{w}_{\rm FMA}}}(x)-f_{\rm WENS}(x)\|_2)
\\= & \mathcal{L} ({ \mathbf{\{w}_k\}_{k=1}^K})  +  O(\Delta^2 ).
    \end{aligned}
    \label{decomposistion:step7}
\end{equation}

Finally, combining (\ref{decomposistion:step4}) and (\ref{decomposistion:step6}) with (\ref{decomposistion:step2}), we have:
\begin{equation}
    \begin{aligned}
 & \mathbb{E}_{ \mathbf{\{w}_k\}_{k=1}^K \in \prod_{k}^K \mathcal{W}_{\mathcal{D}_{k}}}  \mathcal{L} (\mathbf{{w}_{\rm FMA}} ) 
 \\ = &   \mathbb{E}_{ \mathbf{\{w}_k\}_{k=1}^K \in \prod_{k}^K \mathcal{W}_{\mathcal{D}_{k}}} \mathcal{L} ({ \mathbf{\{w}_k\}_{k=1}^K}) +  O(\Delta^2 )\\
 \\ = &   \mathbb{E}_{(x,y) \in \mathcal{D}} [( {\rm{Bias}}  \{f_{\rm WENS}|(x,y)\})^2 + {\rm{Var}}  \{f_{\rm WENS}|x\}  ] +  O(\Delta^2 )\\
 \\ = &  \frac{1}{n} \sum_{(x,y)\in\mathcal{D}}
 [\sum_{k=1}^K \frac{n_k}{n}  {\rm{TrainBias}} \{f_{\mathbf{w}_k}|(x,y)\} 
  \\   & +  \frac{n_k}{n} {\rm{HeterBias}} \{f_{\mathbf{w}_k}|(x,y)\} ]^2
  \\   & +   { \sum_{k=1}^K \frac{n_k^2}{n^2} {\rm Var}\{f_{\mathbf{w}_k}|x\}} 
  +   {\sum_k \sum_{{k^\prime}} \frac{n_k n_{{k^\prime}}}{n^2}  {\rm Cov}\{f_{\mathbf{w}_k,\mathbf{w}_{k^\prime}}|x\}}  \\
  \\   &  +  O(\Delta^2 ).
    \end{aligned}
    \label{decomposistion:step8}
\end{equation}
 
\section*{Appendix B:  Loss Landscape Visualization}
\subsection*{Loss Landscape Visualization of Cross-device and Cross-silo FL}
\begin{figure*}[t]
    \centering
    \includegraphics[width=\textwidth]{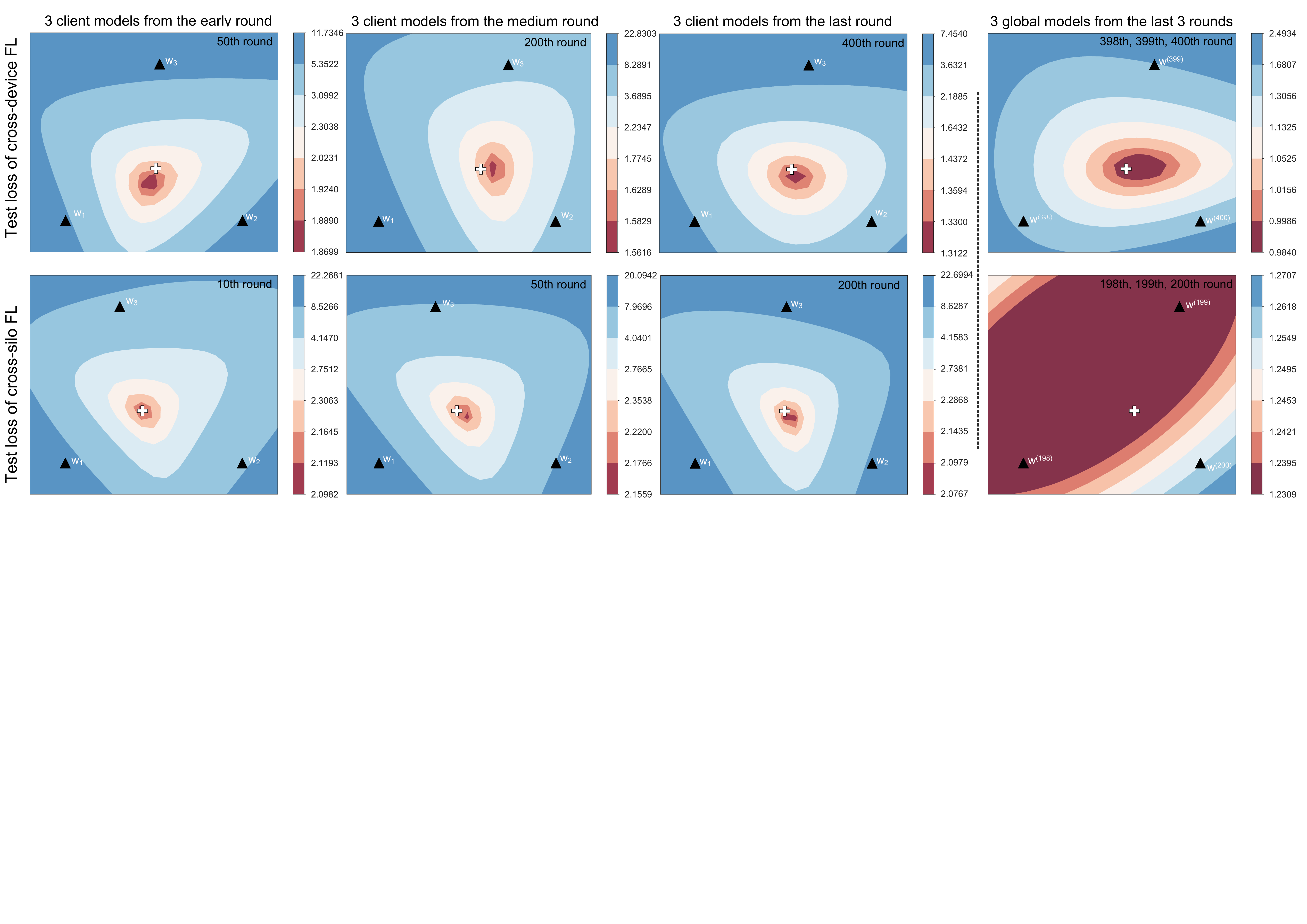}
    \caption{Loss landscape visualization with three models on global test set in cross-device and cross-silo FL.}
    \label{loss_visualization_cross_silo_FL}
\end{figure*}

We examine two common FL frameworks to demonstrate the similarity of loss landscapes across different FL frameworks: cross-device FL and cross-silo FL \cite{kairouz2021advances}.
The number of clients involved in cross-silo FL is small (e.g., the cross-silo FL shown in Figure \ref{loss_visualization_cross_silo_FL} includes ten clients), and all clients participate fully in each communication round.
On the other hand, cross-device FL requires a large number of clients, and only a subset of them participate in each round (e.g., the cross-device FL  in  Figure \ref{loss_visualization_cross_silo_FL} involves 100 clients with 0.1 participation rate for each round).
Figure \ref{loss_visualization_cross_silo_FL} depicts the loss landscape visualization with three models on the global dataset for both cross-device and cross-silo FL settings.

Similar to the  FMA's geometric properties observed from Figure \ref{loss_error_visualization}, the FMA model (i.e.,  the  \textit{white cross}) achieves lower test loss and error than the individual client models (i.e.,  the \textit{black triangles}) throughout the training process in both settings.
Furthermore,   both FL settings illustrate that FMA maintains the client and global models closely within a shared basin.
Notably, the deviation between the \textit{white cross} and the lowest point of the basin in terms of loss/error is smaller in cross-device FL than cross-silo FL, as shown in Figure \ref{loss_visualization_cross_silo_FL}. 
This finding supports the analysis presented in Section \ref{section:proposed_method}, which suggests that low participation rates exacerbate the deviation.

 \begin{figure*}[th]
    \centering
    \includegraphics[width=\textwidth]{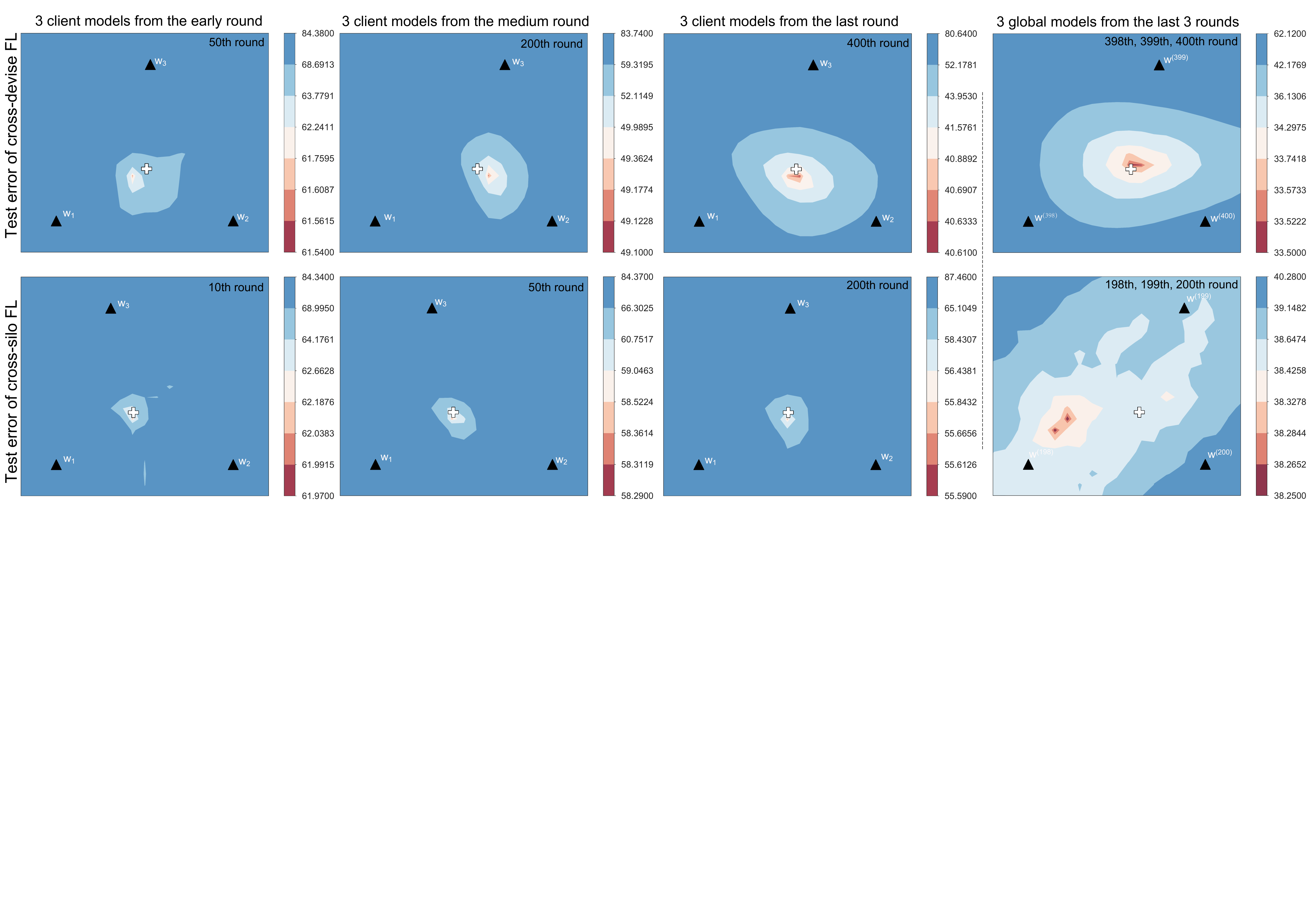}
    \caption{Error landscape visualization with three models on global test set  in cross-device and cross-silo FL.}
    \label{error_landscape}
\end{figure*}

In addition to the loss landscape,  we visualize the classification error landscape on the global dataset for both settings in Figure \ref{error_landscape}. 
The observed geometric properties of FMA  in Figure \ref{error_landscape} are similar to those in  Figure \ref{loss_visualization_cross_silo_FL}. 
Therefore, we omit the detailed descriptions here to avoid repetition.

\subsection*{Loss Landscape Visualization under Different Models, Datasets, and Heterogeneous Data Settings}
\begin{figure*}[t]
    \centering
    \includegraphics[width=\textwidth]{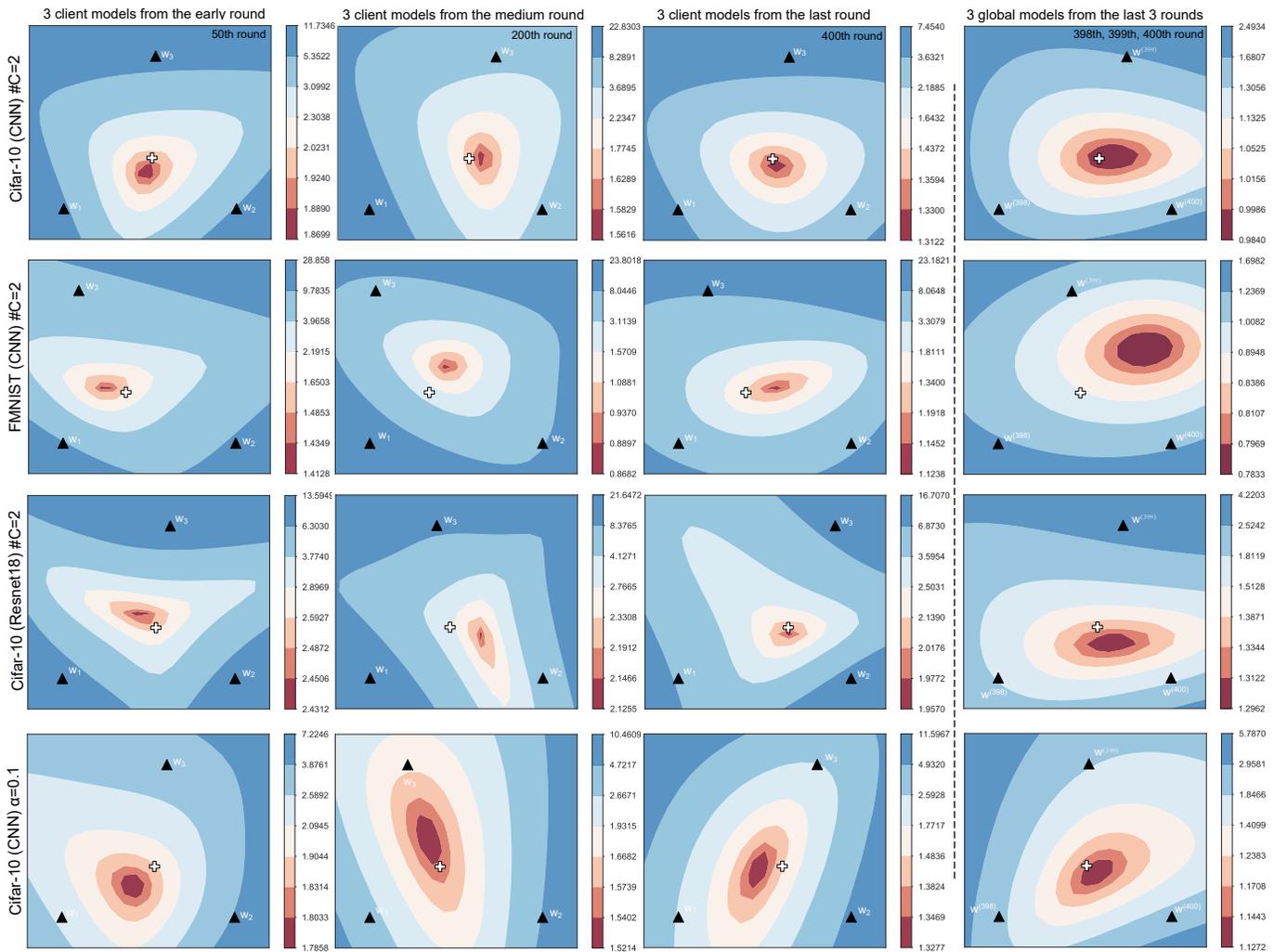}
    \caption{
    Loss landscape visualization of client and global models in the early, medium, and final stages of FL  under different setups:
    \textbf{The first row} illustrates the visualization of  CNN models trained under data heterogeneity $\#C=2$ on CIFAR-10;
    \textbf{The second row} illustrates the visualization of  CNN models trained under $\#C=2$ on FMNIST;
    \textbf{The third row} illustrates the visualization of  ResNet18 trained  under  $\#C=2$ on CIFAR-10;
    \textbf{The fourth row} illustrates the visualization of  CNN models trained with a Dirichlet parameter $\alpha=0.1$ on CIFAR-10.  
    Each row shows the loss landscape visualization for local and global models in the early, medium, and final stages.}
    \label{full_loss_landscape}
\end{figure*}

To further explore the geometric properties of  FMA, we visualize the loss landscape of  FL under various models (including the CNN model and the ResNet model), datasets (including FMNIST and CIFAR-10), and data heterogeneity (including label skews with $\#C=2$ and $\alpha=0.1$).
The visualization results are presented in Figure \ref{full_loss_landscape}.
The geometric properties of FMA discussed in Section \ref{section:loss_landscape} are consistent with those observed in Figure \ref{full_loss_landscape}.
 Regardless of the specific FL setup, FMA ensures that client and global models reside within a common basin.
This geometric insight sheds light on how FMA  effectively prevents client models from over-fitting to their respective datasets (i.e., FMA mitigates the over-fitting information of client models being aggregated into the global model) and improves the generalization performance of the global model.

\section*{Appendix C: Further Analysis}\label{Further_analysis_var}
To analyze  the variance term, we set $n_i = n_j$ (i.e., the number of client samples is the same) to isolate the impact of weighted averaging on the loss decomposition in Theorem \ref{theorem1}. Consequently, we have the following corollary:
\begin{corollary}
 (Loss decomposition of FMA with the same client sample sizes. Extended from Theorem \ref{theorem1}.) 
Given $K$ client models $ \mathbf{\{w}_k\}_{k=1}^K \in \prod_{k}^K \mathcal{W}_{\mathcal{D}_{k}}$ and $n_i/n = n_j/n = 1/K$, we can decompose the expected loss of the FMA's model $\mathbf{w}_{\rm FMA}$ on  $\mathcal{D}$ as:
\begin{equation}
    \begin{aligned}
\mathbb{E}_{ \mathbf{\{w}_k\}_{k=1}^K}  \mathcal{L} (\mathbf{{w}_{\rm FMA}} )
   = \frac{1}{nK^2} \sum_{(x,y)\in\mathcal{D}}
   [\sum_{k=1}^K  {\rm{TrainBias}} \{f_{\mathbf{w}_k}|(x,y)\}  
 \\  +  {\rm{HeterBias}} \{f_{\mathbf{w}_k}|(x,y)\}  ]^2  
   +  \frac{1}{K} \underbrace{ \sum_{k=1}^K \frac{1}{K} {\rm Var}\{f_{\mathbf{w}_k}|x\}}_{\overline{\rm Var}\{f_{ \mathbf{\{w}_k\}_{k=1}^K}|x\}} 
 \\  +    \sum_k \frac{K-1}{K^2}  \underbrace{\sum_{{k^\prime}\neq k} \frac{1}{K-1} {\rm Cov}\{f_{\mathbf{w}_k,\mathbf{w}_{k^\prime}}|x\}}_{\overline{\rm Cov}\{f_{ \mathbf{{w}_k},\mathbf{{w}_{k^\prime}}}|x\}}   
 +  O(\Delta^2 ),
    \end{aligned}
\end{equation}
where ${\rm Var}\{f_{\mathbf{w}_k}|x\} =  \mathbb{E}_{ \mathbf{{w}_k} }
     [ (
    f_{\mathbf{w}_k}(x) - \mathbb{E}_{ \mathbf{{w}_k} }[f_{\mathbf{w}_k}(x)]
     )^2 ]$;
${\rm Cov}\{f_{\mathbf{w}_k,\mathbf{w}_{k^\prime}}|x\} =  \mathbb{E}_{ \mathbf{{w}_k},\mathbf{{w}_{{k^\prime}}}}
     [  (f_{\mathbf{w}_k}(x) - \mathbb{E}_{ \mathbf{{w}_k} }[f_{\mathbf{w}_k}(x)] ) $$  (f_{\mathbf{w}_{{k^\prime}}}(x) $ $ - \mathbb{E}_{ \mathbf{{w}_{{k^\prime}}} }[f_{\mathbf{w}_{{k^\prime}}}(x)]) ]$; 
    $\overline{\rm Var}\{f_{ \mathbf{\{w}_k\}_{k=1}^K}|x\}$ denotes the mean variance of client models when given a sample $(x,y) \in \mathcal{D}$; 
    $\overline{\rm Cov}\{f_{ \mathbf{{w}_k},\mathbf{{w}_{k^\prime}}}|x\} $ denotes the mean covariance between two client models when given a sample $(x,y) \in \mathcal{D}$.
\label{corollary2}
\end{corollary}
With Corollary \ref{corollary2}, the mean covariance   $\overline{\rm Cov}\{f_{ \mathbf{{w}_k},\mathbf{{w}_{k^\prime}}}|x\} $  and  the mean variance  $\overline{\rm Var}\{f_{ \mathbf{\{w}_k\}_{k=1}^K}|x\}$ become equivalent  when client models are identically distributed
(i.e., client models are trained on homogeneous datasets with the same training configurations).
In this case, $\overline{\rm Var}\{f_{ \mathbf{\{w}_k\}_{k=1}^K}|x\} = \overline{\rm Cov}\{f_{ \mathbf{{w}_k},\mathbf{{w}_{k^\prime}}}|x\} = {\rm Var}\{f_{ \mathbf{w}_k}|x\}$, and thus the effect of the variance and covariance factors in Theorem \ref{theorem1} is the same as that of a single client model, making the aggregation of more client models in FMA useless.

\section*{Appendix D: Experiment Settings}\label{appendix:experiment_setups}
\subsection*{Models}
Table \ref{table:specific_model} outlines the models used in all the experiments, including validation, test, and ablation experiments. 
To isolate the controversial effect of BN layers on FL, we follow \cite{hsieh2020non} to replace the BN layer with the GroupNorm layer in all experiments. 
Our models adhere to the architectures reported in the respective baseline works for a fair comparison. 
Specifically, for the experiments conducted on the FMNIST and CIFAR-10 datasets, we employ a CNN model consisting of two 5x5 convolutional layers followed by 2x2 max pooling and two fully connected layers with ReLU activation.
This architecture aligns with the model used in \cite{mcmahan2017communication,zhou2022fedfa}.
For the CIFAR-10/100 experiments, we adopt the ResNet-18 architecture \cite{he2016deep} with a linear projector. This choice is consistent with the models employed in \cite{reddi2021adaptive} and \cite{li2021fedbn}.
Lastly, for the Digit Fives dataset, we employ a CNN model with three 5x5 convolutional layers followed by five GroupNorm layers.

\subsection*{Baseline Settings}
 Table \ref{table:hyperparameter} provides the additional hyper-parameters specific to different baselines. 
These hyper-parameters are chosen based on the setups reported in the respective baseline works.
Here is a brief description of the role of hyper-parameters in each baseline:
\begin{itemize}
    \item FedProx and FedFA: These baselines modify the loss function by adding a proximal term at the client side.
The best coefficient of the proximal term is selected from the given range [0.1, 0.01, 0.001]. 
This coefficient controls the trade-off between the proximal and main loss functions.
    \item FedASAM: This baseline utilizes the SAM technique as the client loss function. 
    The hyper-parameters $\eta_{\rm SAM}$ and $\rho_{\rm SAM}$ control the noise introduced in SAM, affecting the exploration-exploitation trade-off during optimization.
    \item FedAdam and FedYogi: These baselines apply adaptive momentum to the global update on the server side.  
    The hyper-parameters include the server learning rate (lr) $\eta$, decay parameters $\beta_1$, $\beta_2$, and the degree of adaptivity $\tau_1$.
    These hyper-parameters control the adaptation of the server-side optimizer's momentum over time.
    \item  FedGMA: This baseline employs the AND-Masked gradient update based on the masking threshold $\epsilon$. 
    The masking threshold determines the sparsity level in the gradient updates and improves the flatness of the global model. 
\end{itemize}
 It is noteworthy that FedAvg and FedNova do not require additional hyper-parameters beyond the standard optimization parameters.
\begin{table*}[th]
\centering
\caption{
Parameter settings for all the models used in our experiments. Group normalization (GN) layers split the input channels into two groups in all models.
Con2d($a,b,c$) represents a convolutional layer with $a$ input channels, $b$ output channels, and $c\times c$ kernel sizes.
FC($a,b$) denotes a fully connected (FC) layer with $a$ input channels and $b$ output channels. 
MaxPool2D($a,b$) is a max pooling layer with dimensions $a\times b$, and ReLU refers to the ReLU activation function.
The backbone refers to the framework excluding the last layer. For example, in ResNet18, the backbone corresponds to its feature extractor. ResNet18.FC($a,b$) represents ResNet18 with the classifier replaced by an FC($a,b$) layer.
}
\label{table:specific_model}
\resizebox{\textwidth}{!}{%
\begin{tabular}{@{}cc|ccc|c|c@{}}
\toprule
\multirow{2}{*}{} & \multicolumn{6}{c}{Dataset (Used Model)} \\ \cmidrule(l){2-7} 
 & FMNIST & \multicolumn{3}{c}{CIFAR-10/100} & Digit Five & PACS \\
Block & (CNN) & (CNN) & (VGG11) & (ResNet18) & (CNN) & (AlexNet) \\ \midrule
1 & \begin{tabular}[c]{@{}c@{}}Conv2d(1,32,5)\\  ReLU,MaxPool2D(2,2)\end{tabular} & \begin{tabular}[c]{@{}c@{}}Conv2d(3,64,5)\\  ReLU,MaxPool2D(2,2)\end{tabular} & \begin{tabular}[c]{@{}c@{}}Backbone of\\  VGG11 \\ with GN\end{tabular} & \begin{tabular}[c]{@{}c@{}}Backbone of \\ ResNet18\\ with GN\end{tabular} & \begin{tabular}[c]{@{}c@{}}Conv2d(3,64,5,1,2)\\ GN(2,64)\\  ReLU,MaxPool2D(2,2)\end{tabular} & \begin{tabular}[c]{@{}c@{}}Backbone of \\ AlexNet\\ with GN\end{tabular} \\
2 & \begin{tabular}[c]{@{}c@{}}Conv2d(1,32,5)\\  ReLU,MaxPool2D(2,2)\end{tabular} & \begin{tabular}[c]{@{}c@{}}Conv2d(64,64,5)\\ ReLU,MaxPool2D(2,2)\end{tabular} & VGG11.FC(512,128) & ResNet18.FC(512,128) & \begin{tabular}[c]{@{}c@{}}Conv2d(64,64,5,1,2)\\ GN(2,64)\\  ReLU,MaxPool2D(2,2)\end{tabular} & ResNet18.FC(4096,512) \\
3 & \begin{tabular}[c]{@{}c@{}}FC(512,384)\\ ReLU\end{tabular} & \begin{tabular}[c]{@{}c@{}}FC(1600,384)\\ ReLU\end{tabular} & FC(128,10) & FC(128,10) & \begin{tabular}[c]{@{}c@{}}Conv2d(64,128,5,1,2)\\ GN(2,128)\\  ReLU,MaxPool2D(2,2)\end{tabular} & FC(512,10) \\
4 & FC(384,128) & FC(384,128) &  &  & \begin{tabular}[c]{@{}c@{}}FC(6272, 2048)\\ ReLU\end{tabular} &  \\
5 & FC(128,10) & FC(128,10) &  &  & FC(2048,128) &  \\
6 &  &  &  &  & FC(128,10) &  \\ \bottomrule
\end{tabular}%
}
\end{table*}

\begin{table*}[t]
\centering
\caption{
Hyper-parameter setups for all the baselines. FedAvg and FedNova are excluded as they do not require additional hyper-parameters.
In FedASAM, $\eta_{\rm SAM}$ and $\rho_{\rm SAM}$ control the noise to affect the exploration-exploitation trade-off during optimization.
For FedAdam and FedYogi,   the server learning rate $\eta$, momentum decay parameters $\beta_1$, $\beta_2$, and the adaptivity degree $\tau_1$ control the adaptation of the server-side optimizer's momentum over time.
In FedGMA, the masking threshold $\epsilon$   determines the sparsity level in the gradient updates to improve the flatness of the global model loss. 
Our code for these baselines follows the hyperlinks provided below.}
\label{table:hyperparameter}
\resizebox{\textwidth}{!}{%
\begin{tabular}{@{}ccccccc@{}}
\toprule
\begin{tabular}[c]{@{}c@{}}Hyper-\\ parameter\end{tabular} & \begin{tabular}[c]{@{}c@{}}FedProx\\  \end{tabular} & \begin{tabular}[c]{@{}c@{}}FedASAM\\  \end{tabular} & \begin{tabular}[c]{@{}c@{}}FedFA\\ \end{tabular} & \begin{tabular}[c]{@{}c@{}}FedAdam\\ \end{tabular} & \begin{tabular}[c]{@{}c@{}}FedYogi\\ \end{tabular} & \begin{tabular}[c]{@{}c@{}}FedGMA\\ \end{tabular} \\ \midrule
 & \begin{tabular}[c]{@{}c@{}}coefficient of proximal term:\\ Best from [0.1, 0.01,0.001]\end{tabular} & \begin{tabular}[c]{@{}c@{}}CNN models: $\rho_{\rm SAM}$=0.7,\\  $\eta_{\rm SAM}$=0.2\end{tabular} & \begin{tabular}[c]{@{}c@{}}coefficient of proximal term:\\ Best from [0.1, 0.01,0.001]\end{tabular} & \multicolumn{2}{c}{$\eta$ = 0.01} & \begin{tabular}[c]{@{}c@{}}   $\epsilon = 0.8$\end{tabular} \\
 &  & \begin{tabular}[c]{@{}c@{}}Other models: $\rho_{\rm SAM}$=0.2,\\  $\eta_{\rm SAM}$=0.05\end{tabular} & \begin{tabular}[c]{@{}c@{}}coefficient of anchor \\  updates: 0.9\end{tabular} & \multicolumn{2}{c}{$\beta_1$  = 0.9} & \begin{tabular}[c]{@{}c@{}}  \end{tabular} \\
 &  &  &  & \multicolumn{2}{c}{$\beta_2$ = 0.99} &  \\
 &  &  &  & \multicolumn{2}{c}{$\tau_1$ = 0.001} &  \\
Refer & - & \href{https://github.com/debcaldarola/fedsam}{Authors' codes} & 
\href{https://github.com/TailinZhou/FedFA}{Authors' codes}& \multicolumn{2}{c}{\href{https://github.com/adap/flower/blob/933e4d8dea15c948ac11810d487e9a77da9b45bf/src/py/flwr/server/strategy/fedyogi.py}{Benchmark:Flower}} & \href{https://github.com/siddarth-c/FedGMA}{Reproduces codes} \\ \bottomrule
\end{tabular}%
}
\end{table*}
\subsection*{Settings of  Visualization and Validation experiments}\label{appendix:Setup of all visualization}
Table \ref{Setup_all_visuz_exps} provides the specific setups for all visualization and validation experiments. 
The experiments are performed using PyTorch on a single node of the High-Performance Computing platform. The node has 4 NVIDIA A30 Tensor Core GPUs, each with 24GB of memory. 
The setups include hyper-parameters and configurations specific to each experiment, such as the model architecture, dataset, number of clients, batch size, learning rate, optimizer, and other relevant details.
\begin{table*}[t]
\centering
\caption{
Setup of all visualization and validation experiments in this work. 
Each row details the specific experiment setup corresponding to the figures.
Cross-device FL and cross-silo FL indicate that some and all clients participate in each training round, respectively.
$\#C=2$ implies that each client holds two class shards of the training dataset, with each shard containing 250 samples. Moreover, $\alpha=0.1$ represents the splitting of the training dataset using a Dirichlet distribution $Dir(\alpha)=0.1$ as in \cite{yurochkin2019bayesian}.
}
\label{Setup_all_visuz_exps}
\resizebox{\textwidth}{!}{%
\begin{tabular}{@{}cccccccccccc@{}}
\toprule
 & FL & \begin{tabular}[c]{@{}c@{}}client \\ number\end{tabular} & \begin{tabular}[c]{@{}c@{}}client \\ participation\end{tabular} & \begin{tabular}[c]{@{}c@{}}local \\ epoch\end{tabular} & \begin{tabular}[c]{@{}c@{}}local\\ batch\end{tabular} & \begin{tabular}[c]{@{}c@{}}lr\\ (momentum)\end{tabular} & \begin{tabular}[c]{@{}c@{}}lr decay\\ per round\end{tabular} & round & dataset & \begin{tabular}[c]{@{}c@{}}heterogeneious\\ data(C:class)\end{tabular} & model \\ \midrule
Figure \ref{loss_error_visualization}, \ref{error_landscape} & Cross device & 100 & 0.1 & 5 & 50 & 0.01(0.9) & 0 & 400 & CIFAR-10 & $\#C=2$ & CNN \\
Figure \ref{loss_visualization_cross_silo_FL}, \ref{error_landscape} & Cross silo  & 10 & 1 & 5 & 50 & 0.01(0.9) & 0 & 200 & CIFAR-10 & $\#C=2$ & CNN \\
Figure \ref{full_loss_landscape} & Cross device & 100 & 0.1 & 5 & 50 & 0.01(0.9) & 0 & 400 & CIFAR-10, FMNIST & $\#C=2$,$\alpha=0.1$ & CNN, ResNet18 \\
Figure \ref{Bias_validation}, \ref{locality_validation},\ref{Variance_covariance_validation} & Cross silo & 10 & 1 & 5 & 50 & 0.01(0.9) & 0/0.01 & 400 & CIFAR-10 & $\#C=2$ & CNN \\
Figure \ref{IMA_validation} & Cross device & 100 & 0.1 & 5 & 50 & 0.01(0.9) & 0 & 400 & CIFAR-10 & $\#C=2$ & CNN \\ \bottomrule
\end{tabular}%
}
\end{table*}

\subsection*{Settings of Test Experiments}\label{Setup of test experiments}
Table \ref{table:testsetup} provides the setup details for all test experiments conducted in this work.
The table includes information such as the client number, participation rate, local epoch number, lr, decay scheme, total communication rounds, model, and specific setups of IMA.
 Please refer to Table \ref{table:testsetup} for a comprehensive overview of the experimental configurations. 
 \begin{table*}[t]
\centering
\caption{
Setup of all test experiments in Table \ref{table:performance}. 
Each row shows the specific experiment setup for the corresponding datasets, including the FL, IMA, and tested models.
For the IMA setup, the columns "IMA windows" and "starting IMA" denote $P$ and $t_s = 0.75 R$ in (\ref{iterative model averaging}), respectively.
}
\label{table:testsetup}
\resizebox{\textwidth}{!}{%
\begin{tabular}{@{}cccccccccccc@{}}
\toprule
Dataset & \begin{tabular}[c]{@{}c@{}}client \\ number\end{tabular} & \begin{tabular}[c]{@{}c@{}}client \\ participation\end{tabular} & \begin{tabular}[c]{@{}c@{}}local \\ epoch\end{tabular} & \begin{tabular}[c]{@{}c@{}}local\\ batch\end{tabular} & \begin{tabular}[c]{@{}c@{}}lr\\ (momentum)\end{tabular} & \begin{tabular}[c]{@{}c@{}}lr decay\\ per round\end{tabular} & round ($R$) & \begin{tabular}[c]{@{}c@{}}IMA\\ windows\end{tabular} & \begin{tabular}[c]{@{}c@{}}IMA lr\\ decay\end{tabular} & \begin{tabular}[c]{@{}c@{}}starting\\ IMA   \end{tabular} & model \\ \midrule
FMNIST & 100 & 0.1 & 5 & 50 & 0.01(0.9) & 0.01 & 300 & 5 & 0.03 & 225 & CNN \\
CIFAR-10 & 100 & 0.1 & 5 & 50 & 0.01(0.9) & 0.01 & 400,400 & 5 & 0.03 & 300,300 & CNN,ResNet18 \\
CIFAR-100 & 100 & 0.1 & 5 & 50 & 0.01(0.9) & 0.01 & 300,400 & 5 & 0.03 & 225,300 & VGG11,ResNet18 \\
Digit Five & 100 & 0.1 & 5 & 50 & 0.01(0.9) & 0.01 & 200 & 5 & 0.03 & 150 & CNN w/ GN \\
PACS & 80 & 0.2 & 5 & 50 & 0.01(0.9) & 0.01 & 400 & 5 & 0.03 & 300 & AlexNet w/ GN \\ \bottomrule
\end{tabular}%
}
\end{table*}

\end{document}